\documentclass[11pt]{article}

\usepackage[preprint]{acl}

\usepackage{times}
\usepackage{latexsym}

\usepackage[T1]{fontenc}

\usepackage[utf8]{inputenc}

\usepackage{microtype}

\usepackage{inconsolata}

\usepackage{graphicx}

\usepackage{graphicx}
\usepackage{subcaption}
\usepackage{booktabs}
\usepackage{comment}
\usepackage{amsmath}
\usepackage{amssymb}
\usepackage{multirow}
\usepackage{CJKutf8}
\newcommand{\zh}[1]{\begin{CJK*}{UTF8}{gbsn}#1\end{CJK*}}
\newcommand{\jp}[1]{\begin{CJK*}{UTF8}{min}#1\end{CJK*}}  
\newcommand{\ko}[1]{\begin{CJK*}{UTF8}{mj}#1\end{CJK*}}  

\usepackage{tcolorbox}
\tcbset{
  examplebox/.style={
    colback=gray!5,
    colframe=gray!50,
    boxrule=0.5pt,
    arc=3pt,
    left=6pt,
    right=6pt,
    top=4pt,
    bottom=4pt
  }
}

\title{Generation-Step-Aware Framework for Cross-Modal Representation and Control in Multilingual Speech–Text Models}

\author{
  Toshiki Nakai$^{1}$\thanks{This work was completed while the author was at Saarland University.},
  Varsha Suresh$^{2}$,
  Vera Demberg$^{2,3}$ \\[2mm]
  $^{1}$Leipzig University, ScaDS.AI Dresden/Leipzig \\
  $^{2}$Saarland University \\
  $^{3}$Max Planck Institute for Informatics, Saarland Informatics Campus \\
  \texttt{toshiki3738@gmail.com}
}

\begin{document}
\maketitle
\begin{abstract}
Multilingual speech–text models rely on cross-modal language alignment to transfer knowledge between speech and text, but it remains unclear whether this reflects shared computation for the same language or modality-specific processing. We introduce a generation-step-aware framework for evaluating cross-modal computation that (i) identifies language-selective neurons for each modality at different decoding steps, (ii) decomposes them into language-representation and language-control roles, and (iii) enables cross-modal comparison via overlap measures and causal intervention. Applying our framework to SeamlessM4T, and additionally evaluating its generality on the decoder-only model Qwen2-Audio, we find that cross-modal language-representation alignment is highly model dependent. SeamlessM4T exhibits pronounced generation-step-dependent specialization, where only 5–7\% of language-representation neurons are shared across modalities and overlap shifts from same-language to typologically related languages during autoregressive generation, whereas Qwen2-Audio maintains substantially larger cross-modal sharing (45–47\%) together with stable, strongly language-specific alignment across decoding steps. In contrast, language-control neurons identified at later decoding steps exhibit progressively stronger cross-modal transfer from speech to text in SeamlessM4T. Together, these results demonstrate that generation-step-aware analysis can reveal both shared and model-specific patterns of cross-modal computation.
\end{abstract}

\noindent\textbf{Code:} \url{https://github.com/konta3738/generation-step-aware-framework}

\section{Introduction}

Multilingual speech--text foundation models aim to unify tasks such as automatic speech recognition, speech translation, and text translation within a single architecture that processes both spoken and written language \citep{10.5555/3618408.3618627, rubenstein2023audiopalmlargelanguagemodel, tang2024salmonngenerichearingabilities}. A central premise behind such models is \emph{cross-modal language alignment}: representations associated with the same language should remain sufficiently compatible across speech and text for multilingual knowledge to transfer across modalities. If this alignment is weak, models may rely on modality-specific computation even for the same language, limiting generalization.

Despite its importance, it remains unclear whether the same circuitry is reused to represent or output the same language across different modalities, and how. Prior mechanistic interpretability work on multilingual \emph{text-only} models identifies language-selective neurons which react specifically to certain languages, finding that those are concentrated in early and late layers, with mid-layers being language-agnostic \citep{kojima-etal-2024-multilingual, tang-etal-2024-language, tan-etal-2024-neuron, gurgurov-etal-2025-language, trinley2025languagesdoesaya23think}. However, comparable neuron-level evidence for \emph{multimodal} models remains scarce. In particular, we lack a clear account of whether multilingual speech--text models have language-selective neurons that react to both speech and text inputs, or instead rely on modality-specific processing, without shared mechanism for the same language.

A key limitation of existing neuron-based methods \citep{cuadros2022self, kojima-etal-2024-multilingual} is that they collect decoder activations under full-sequence conditioning, rather than the prefix available at each decoding step. This mismatches the autoregressive computation, where predictions are conditioned only on previously generated tokens and therefore vary across steps. In encoder-decoder speech models, early decoding relies primarily on encoder-derived representations, whereas later steps increasingly depend on target-side context \citep{Reid2023InterpretingWhisper}. This issue is particularly important in multilingual speech--text models, where encoder representations carry modality-specific information.

In this work, we revisit cross-modal language alignment through the lens of \emph{functional decomposition}. We ask: \emph{how and when do multilingual speech--text models reuse shared circuitry across modalities, and how does this reuse differ between input-language representation and output-language generation during decoding?} To answer this, we introduce a generation-step-aware diagnostic framework, which extends prior Average Precision (AP) based neuron identification for text models \citep{cuadros2022self, kojima-etal-2024-multilingual} in three key ways: it (i) identifies language-selective neurons separately for each modality at different decoding steps, (ii) decomposes them by functional role, and (iii) enables direct cross-modal comparison through overlap- and rank-based measures, as well as causal intervention. This unified framework enables systematic evaluation of when, how, and to what extent language processing is shared across modalities. We demonstrate the framework through a detailed case study on SeamlessM4T v2 \citep{communication2023seamlessmultilingualexpressivestreaming}. To examine whether the observed trends generalize beyond encoder–decoder models, we additionally analyze the decoder-only model Qwen2-Audio \citep{chu2024qwen2audiotechnicalreport}.

We decompose decoder-side language processing into two functionally distinct components. \emph{Language-representation neurons} capture how the input language is internally encoded when the output language is fixed, while \emph{language-control neurons} capture mechanisms involved in determining which language is generated. This decomposition enables us to study cross-modal language alignment not as a single property, but as a function of the role that language plays in the computation.

We first analyze SeamlessM4T in detail and then evaluate whether the findings extend to the decoder-only Qwen2-Audio. We reveal two forms of cross-modal language alignment on the input and output sides.

(i) \textbf{Input-side (representation-level alignment).} 
Cross-modal language alignment is highly model dependent. SeamlessM4T exhibits sparse cross-modal neuron sharing (5–7\%) together with timestep-dependent reorganization from same-language to typologically related overlap, whereas Qwen2-Audio exhibits substantially larger sharing (45–47\%) while maintaining stable same-language alignment across decoding steps.

(ii) \textbf{Output-side (control-level alignment).} 
Language-control neurons identified from speech also influence output language in text generation, indicating that output-language control relies on partially shared neuron-level mechanisms across modalities. 
This cross-modal alignment in control becomes stronger at later decoding steps in SeamlessM4T, with the mean cross-modal effect ratio increasing from 0.36 at $t = 6$ to 0.49 at $t = 11$.

Our contributions are threefold. First, we introduce \textbf{generation-step-aware framework}, which models neuron specialization as a function of decoding step and enables cross-modal evaluation of multilingual speech--text models. Second, we characterize \textbf{generation-step-dependent cross-modal language representations} in SeamlessM4T and evaluate whether these trends extend to the decoder-only model Qwen2-Audio. Third, we provide, to the best of our knowledge, \textbf{the first causal evidence of cross-modal steering} at the neuron level, showing that language-control neurons identified from speech transfer to text generation and can be used to change the output language.

\begin{figure*}[t]
  \centering
  \includegraphics[width=\textwidth]{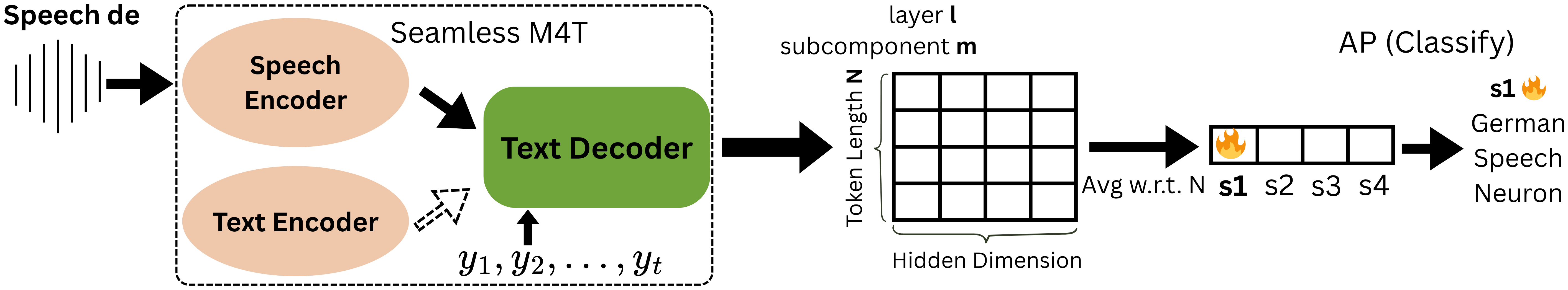}
  \caption{Overview of generation-step-aware neuron identification (length adapter omitted). For each decoding step $t \in {1,6,11}$, we collect activations and identify language-specific neurons via Average Precision (AP) ranking \citep{cuadros2022self, kojima-etal-2024-multilingual}, which are further categorized into \textit{language-representation} and \textit{language-control} neurons.
}
  \label{fig:our_approach}
\end{figure*}

\section{Related work}


\subsection{Language-selective neurons in text-only models}
Prior work identifies language-selective neurons using activation statistics or information-theoretic measures \citep{cuadros2022self, tang-etal-2024-language, tan-etal-2024-neuron, gurgurov-etal-2025-language}, some of which can be used for output-language control. AP-based ranking \citep{kojima-etal-2024-multilingual} selects neurons that best discriminate languages at the full sentence level, ignoring the autoregressive nature of decoding. \citet{tan-etal-2024-neuron} instead identify neurons based on activation frequency in encoder--decoder models, but treat the decoder as effectively static by conditioning on the first decoding step (i.e., BOS token). Existing work is limited to text-only models. Our framework builds on these approaches by explicitly modeling neuron specialization as a function of decoding steps and conditioning context.

\subsection{Mechanistic interpretability in speech--text models}
Mechanistic interpretability for speech and speech--text models remains relatively limited. Prior work on Whisper~\citep{radford2022robustspeechrecognitionlargescale} shows that the encoder primarily captures acoustic structure while the decoder acts as a conditional language model \citep{Reid2023InterpretingWhisper}. Other studies analyze cross-attention contributions \citep{papi2025crossattentionhalfexplanationspeechtotext} and cross-lingual similarity, finding weaker alignment in speech than in text representations \citep{lee-etal-2025-multimodal}. Despite these advances, neuron-level evaluations of multilingual speech--text models are scarce, motivating our study.

\section{Model and method}
\label{sec:method}
This section introduces the experimental setup and diagnostic framework used to study cross-modal processing in multilingual speech--text models. 
We first describe the primary model used in this work, SeamlessM4T v2 \citep{communication2023seamlessmultilingualexpressivestreaming} and the analyzed submodules. 
Next, we present the datasets and tasks used to define language-representation and language-control settings. 
Finally, we introduce our generation-step-aware diagnostic framework.
\subsection{Model and submodules}
Our main experiments use SeamlessM4T v2 Large, which consists of a Conformer-based speech encoder, \citep[w2v-BERT~2.0,][]{communication2023seamlessmultilingualexpressivestreaming}, a Transformer-based text encoder in the style of NLLB, and a shared Transformer text decoder~\citep{nllbteam2022languageleftbehindscaling} with approximately 5B parameters. Each module has 24 layers (0--23). For speech generation, the decoder’s text representations are further processed by additional components (e.g., speech-specific encoders/decoders and a vocoder) to synthesize audio. Since the shared text decoder is the point at which representations from different modalities are unified, we restrict our targets to text outputs and do not examine the speech-generation pathway. We investigate all submodules that expose a hidden (neuron) dimension, including attention, feed-forward, layer normalization, and convolutional components. To evaluate the generality of our findings, we additionally analyze Qwen2-Audio-7B, a decoder-only speech-language model, using the same neuron-identification framework.
\subsection{Data and tasks}
\label{subsec:data_preprocessing}
We use the FLEURS benchmark \citep{conneau2022fleursfewshotlearningevaluation}, a multilingual parallel speech dataset. In this work, we employ FLEURS for four tasks: speech-to-text (S2T) translation, and text-to-text (T2T) translation (X$\rightarrow$En), automatic speech recognition (ASR), and text repetition.

To control for acoustic variability, we synthesize speech from the FLEURS transcriptions using XTTS v2 \citep{casanova2024xttsmassivelymultilingualzeroshot}. This yields 1,000 utterances per language across five languages (German/de, Spanish/es, French/fr, Mandarin/zh, and Japanese/ja). Using single-speaker synthetic speech removes confounds related to speaker identity and recording conditions, allowing us to focus on model-internal representations. 

\subsection{Generation-step-aware diagnostic framework}
\label{subsec:exp1}

We introduce our three-step framework for systematically evaluating cross-modal computation: (i) generation-step-aware neuron identification, (ii) cross-modal comparison, and (iii) cross-modal steering.

\textbf{(i) Generation-step-aware neuron identification.} Our goal is to model neuron specialization as a function of decoding steps and conditioning context, capturing its dynamic nature during autoregressive generation (Figure~\ref{fig:our_approach}). In this step, language-selective neurons are identified separately for each modality.


\paragraph{Neuron definition and pooling.}
Following prior work \citep{cuadros2022self, tan-etal-2024-neuron, tang-etal-2024-language, kojima-etal-2024-multilingual}, we define a \emph{neuron} as a single hidden dimension $j$ of an intermediate activation. Let $X_{\mathrm{enc}}=(x_1,\dots,x_n)$ denote the encoder input and $X_{\mathrm{dec},t}=(y_1,\dots,y_t)$ the decoder prefix available at generation step $t$. For layer $l$ and submodule $m$, let $\mathbf{A}_{\mathrm{enc}}^{(l,m)}(X_{\mathrm{enc}})\in\mathbb{R}^{n\times d}$ and $\mathbf{A}_{\mathrm{dec}}^{(l,m)}(X_{\mathrm{dec},t}\mid X_{\mathrm{enc}})\in\mathbb{R}^{t\times d}$ denote encoder and decoder activations.

For encoder-side modules, we obtain one scalar score per neuron by mean-pooling over the sequence dimension:
\[
\bigl[\mathbf{s}_{\mathrm{enc}}^{(l,m)}\bigr]_j
=
\frac{1}{n}\sum_{i=1}^{n}
\bigl[\mathbf{A}_{\mathrm{enc}}^{(l,m)}(X_{\mathrm{enc}})\bigr]_{i,j}.
\]

For decoder-side modules, in contrast, we define neuron activity at generation step $t$ as the mean activation over all decoder positions generated up to step $t$:
\[
\bigl[\mathbf{s}_{\mathrm{dec}}^{(l,m)}\bigr]_j
=
\frac{1}{t}
\sum_{i=1}^{t}
\bigl[\mathbf{A}_{\mathrm{dec}}^{(l,m)}(X_{\mathrm{dec},t}\mid X_{\mathrm{enc}})\bigr]_{i,j}.
\]

This asymmetric definition matches the model's computation: encoder representations are distributed over the full input sequence, equivalent with the formulation of prior work on decoder-only models \citep{cuadros2022self, kojima-etal-2024-multilingual}, whereas decoder representations are generated autoregressively and are therefore indexed by decoding step. For decoder-only models, the same framework is applied using decoder activations only; the corresponding formulation is provided in Appendix \ref{appdendix:formulation}. 

\paragraph{Average-precision ranking.}
For each layer--submodule pair $(l,m)$ and each decoding condition, we rank neurons by how well their pooled activations discriminate a target language. Let $\mathcal{D}=\{X_{\mathrm{enc},i}, X_{\mathrm{dec},t}\}_{i=1}^{p}$ be a dataset with binary labels $z_i\in\{0,1\}$ indicating membership in a target language class. For neuron $j$, we collect its pooled activations across examples and compute $\mathrm{AP}^{(l,m)}_j = \mathrm{AP}(\mathbf{s}^{(l,m)}_j,\mathbf{z})$,
where higher AP indicates that the neuron more reliably distinguishes the target language. Following \citet{kojima-etal-2024-multilingual}, we select both \textbf{top-$k$} neurons (highest AP) and \textbf{bottom-$k$} neurons (lowest AP), corresponding to excitatory and inhibitory associations with the target language. As baselines, we additionally consider randomly sampled neurons from the same layer--submodule pair (Random) and mid-ranked neurons with AP values near the median (Mid), which serve as non-language-selective neurons.

\paragraph{Label construction.}

We construct binary labels according to the specialization type of interest. In this work, we use \textbf{unimodal language labels}: language classification is performed separately within each modality (speech-only or text-only), so that neuron rankings reflect language distinctions \emph{within} a modality rather than modality differences themselves. This design enables later cross-modal comparison between independently identified language-selective neuron sets.

\paragraph{Types of specialized neurons.}
Having defined how neurons are identified and labeled, we now define two types of language-selective neurons according to the task used to collect activations.

\begin{enumerate}
    \item \textbf{Language-representation neurons.}
    These are identified from speech-to-text translation (S2T) and text-to-text translation (T2T) in the $X \rightarrow \mathrm{En}$ setting. Because the output language is fixed to English while the input language varies, these neurons are intended to capture how input language is internally represented across modalities. 
    In the main analysis, we focus on top-k neurons, as they correspond to the most strongly language-selective units and provide a clearer basis for cross-modal overlap comparison.

    \item \textbf{Language-control neurons.}
    These are identified from automatic speech recognition (ASR), where the output language varies together with the input. They are therefore intended to capture mechanisms involved in output-language generation and control.
    For this intervention, we consider both top-k and bottom-k neurons, since both excitatory and inhibitory associations can influence output-language generation under intervention.
\end{enumerate}

\begin{table*}[t]
\centering
\small
\setlength{\tabcolsep}{5pt}
\renewcommand{\arraystretch}{1.15}

\begin{tabular}{llccccccccc}
\toprule
Model & Metric &
\multicolumn{3}{c}{$t=1$} &
\multicolumn{3}{c}{$t=6$} &
\multicolumn{3}{c}{$t=11$} \\
\cmidrule(lr){3-5}\cmidrule(lr){6-8}\cmidrule(lr){9-11}
&& Random & Mid & Ours &
Random & Mid & Ours &
Random & Mid & Ours\\
\midrule

\multirow{4}{*}{Seamless}
& Cross-modal share
& \textbf{0.14} & 0.05 & 0.05
& \textbf{0.14} & 0.01 & 0.07
& \textbf{0.15} & 0.01 & 0.07 \\

& Same-language
& 0.03 & \textbf{0.81} & 0.46
& -0.00 & 0.03 & \textbf{0.18}
& -0.03 & 0.01 & \textbf{0.21} \\

& Correct typological
& \textbf{0.00} & -0.38 & -0.17
& 0.02 & -0.11 & \textbf{0.19}
& 0.06 & -0.20 & \textbf{0.16} \\

& Wrong typological
& -0.04 & \textbf{-0.75} & -0.48
& -0.03 & 0.10 & \textbf{-0.53}
& -0.05 & 0.25 & \textbf{-0.53} \\

\midrule

\multirow{4}{*}{Qwen2-Audio}
& Cross-modal share
& 0.16 & 0.00 & \textbf{0.45}
& 0.15 & 0.00 & \textbf{0.46}
& 0.14 & 0.00 & \textbf{0.47} \\

& Same-language
& 0.16 & 0.22 & \textbf{1.00}
& 0.15 & 0.14 & \textbf{1.00}
& 0.17 & 0.17 & \textbf{1.00} \\

& Correct typological
& \textbf{-0.03} & -0.29 & -0.47
& -0.05 & \textbf{0.15} & -0.47
& \textbf{-0.06} & -0.10 & -0.47 \\

& Wrong typological
& -0.21 & 0.04 & \textbf{-0.92}
& -0.16 & -0.42 & \textbf{-0.92}
& -0.18 & -0.12 & \textbf{-0.92} \\

\bottomrule
\end{tabular}
\caption{Comparison of overlap metrics. For each model, the three highest values in each row are shown in bold (lowest values for wrong typological).}
\label{tab:overlap_metrics}
\end{table*}

We perform neuron identification at seven decoding steps ($t=1,3,5,6,7,9,11$). In the main paper, we report results for $t=1$, $t=6$, and $t=11$, while results for the remaining decoding steps are provided in Appendix \ref{appendix:additional-steps}. The first step corresponds to generation of the first token, making it the closest setting to previous neuron-identification studies \citep{tan-etal-2024-neuron}, where neuron activations are extracted with only the BOS token provided to the decoder. Later steps reflect decoding after target-side context has begun to accumulate. 

\textbf{(ii) Cross-modal comparison.}
To quantify cross-modal language alignment, we compare neuron sets identified independently from speech and text under matched decoding conditions. For each decoding step, let \(S_i\) denote the top-\(k\) neuron set identified from speech for language \(i\), and \(T_j\) the corresponding top-\(k\) neuron set identified from text for language \(j\), with \(k=1000\). We construct an overlap matrix \(M\) with entries \(M_{ij} = |S_i \cap T_j|\). We evaluate cross-modal alignment using four complementary metrics.

\paragraph{Cross-modal share.}
To measure the overall proportion of language-selective neurons shared across modalities, we compute

\[
C(S,T)=
\frac1L
\sum_i
\frac{|S_i\cap\bigcup_jT_j|}{k},
\]

and compute the symmetric cross-modal share as

\[
\mathrm{CMS}
=
\frac{C(S,T)+C(T,S)}{2},
\]

where \(L\) is the number of languages.

\paragraph{Overlap concentration.}
We partition language pairs into three categories:
same-language (\(i=j\)),
correct typological (\(i\neq j\) within the same typological group),
and wrong typological (different typological groups). For each category \(g\), we compute

\[
R_g=
\frac{\sum_{(i,j)\in g}M_{ij}}
{\sum_{i,j}M_{ij}},
\]

and apply chance correction,

\[
\mathrm{Overlap}_g=
\frac{R_g-q_g}{1-q_g},
\]

where \(q_g\) is the expected overlap ratio under a uniform overlap distribution.

To complement these set-based metrics, we additionally compute Spearman correlation between neuron-wise AP scores across modalities over the full set of neurons (Appendix~\ref{appendix:spearman}).

\textbf{(iii) Cross-modal steering.}
\label{subsec:median}
To test whether identified neurons are causally involved in output-language generation, we perform median-value interventions during inference, following \citep{kojima-etal-2024-multilingual}. Specifically, we replace the activations of targeted neurons with their median values estimated from  corresponding activation distributions.

We apply this intervention to top\&bottom-$1000$ \textbf{language-control neurons} identified from ASR and evaluate whether they also shift the output language in a different modality, namely text repetition. This cross-modal transfer design allows us to test whether output-language control circuitry is shared across speech and text.

We quantify output-language switching using token-level script classification (Latin vs.\ CJK) with regular expressions. Unlike sentence-level language identification tools such as fastText \citep{joulin-etal-2017-bag}, which assume a single dominant label and are unreliable for code-mixed outputs, this provides a simple and robust proxy. We report the \emph{control rate}, defined as the proportion of tokens matching the target script, and use it to assess cross-modal transfer.

We define the \emph{cross-modal effect ratio} as $\mathrm{ControlRate}_B / \mathrm{ControlRate}_A$, where $A$ denotes the source modality used for neuron identification (ASR) and $B$ the target modality (text). This metric quantifies how strongly language-control effects transfer across modalities (details in Appendix \ref{app:ratio}). 

\begin{figure*}[t]
\centering

\begin{tabular}{@{}cccc@{}}
\multicolumn{3}{c}{\textbf{SeamlessM4T}} &
\textbf{Qwen2-Audio}
\\[-2pt]
\cmidrule(lr){1-3}\cmidrule(lr){4-4}
\\[-7pt]

\begin{subfigure}[t]{0.23\linewidth}
    \centering
    \includegraphics[width=\linewidth]{
        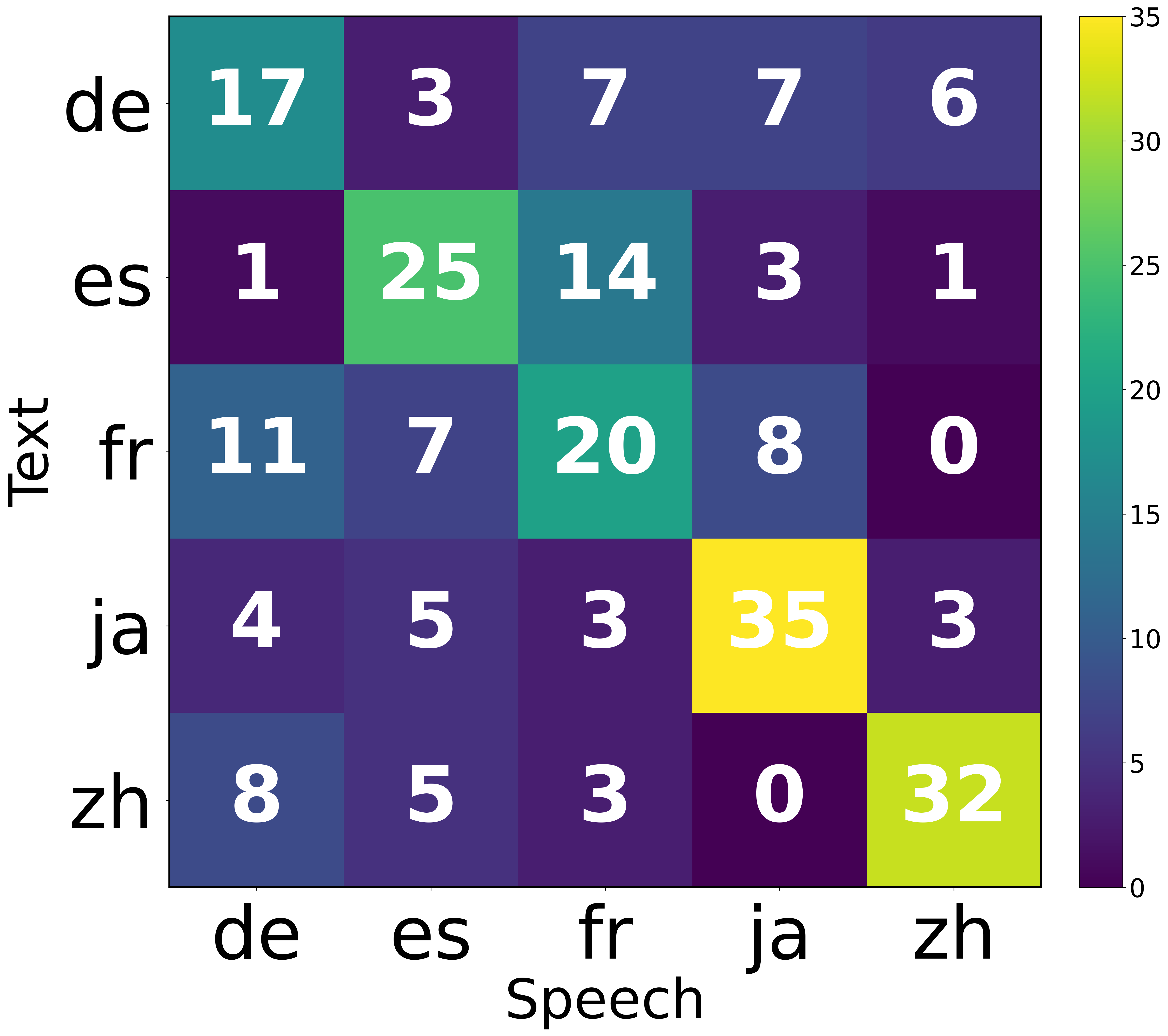
    }
    \caption{Top, $t=1$}
\end{subfigure}
&
\begin{subfigure}[t]{0.23\linewidth}
    \centering
    \includegraphics[width=\linewidth]{
        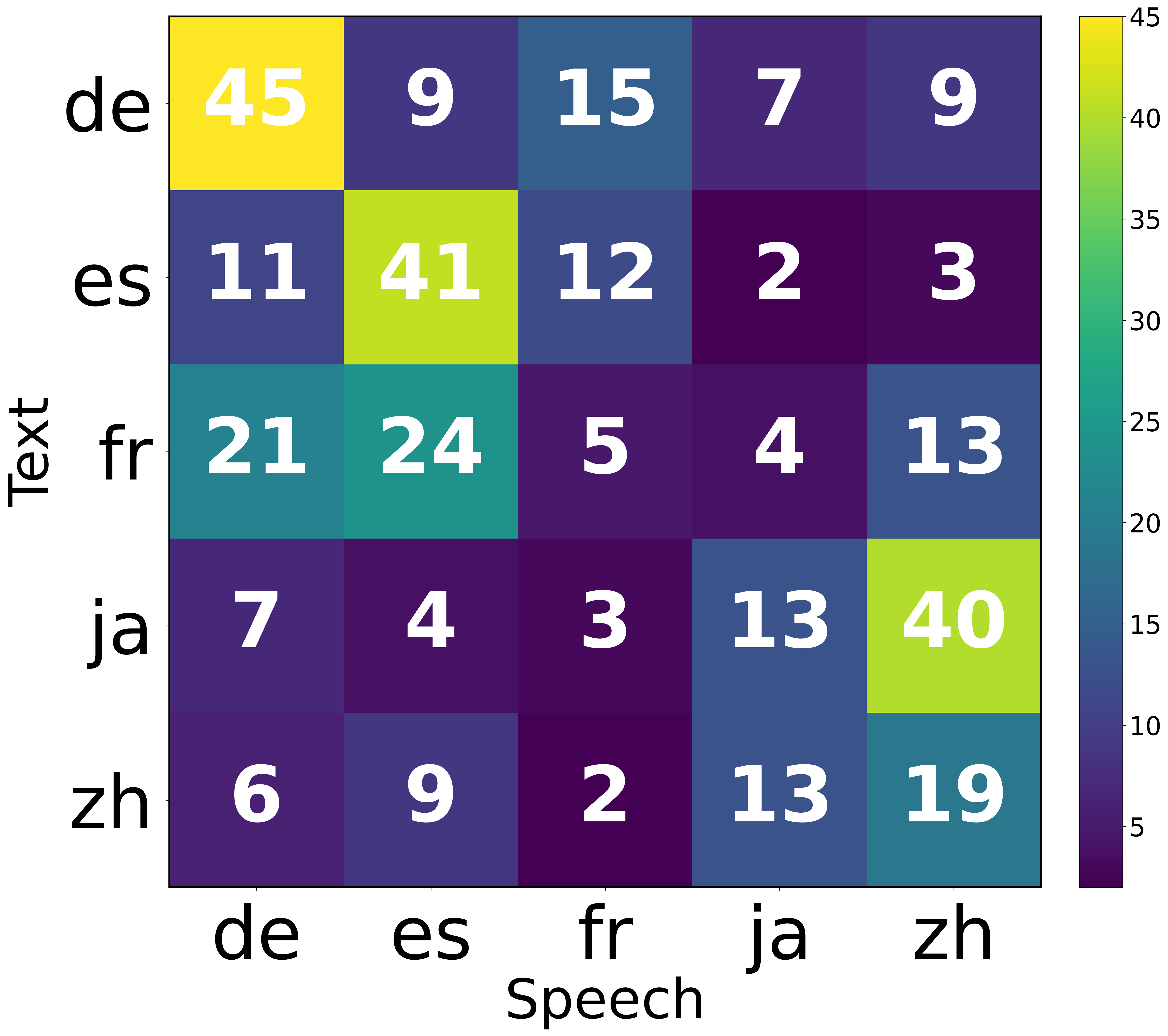
    }
    \caption{Top, $t=11$}
\end{subfigure}
&
\begin{subfigure}[t]{0.23\linewidth}
    \centering
    \includegraphics[width=\linewidth]{
        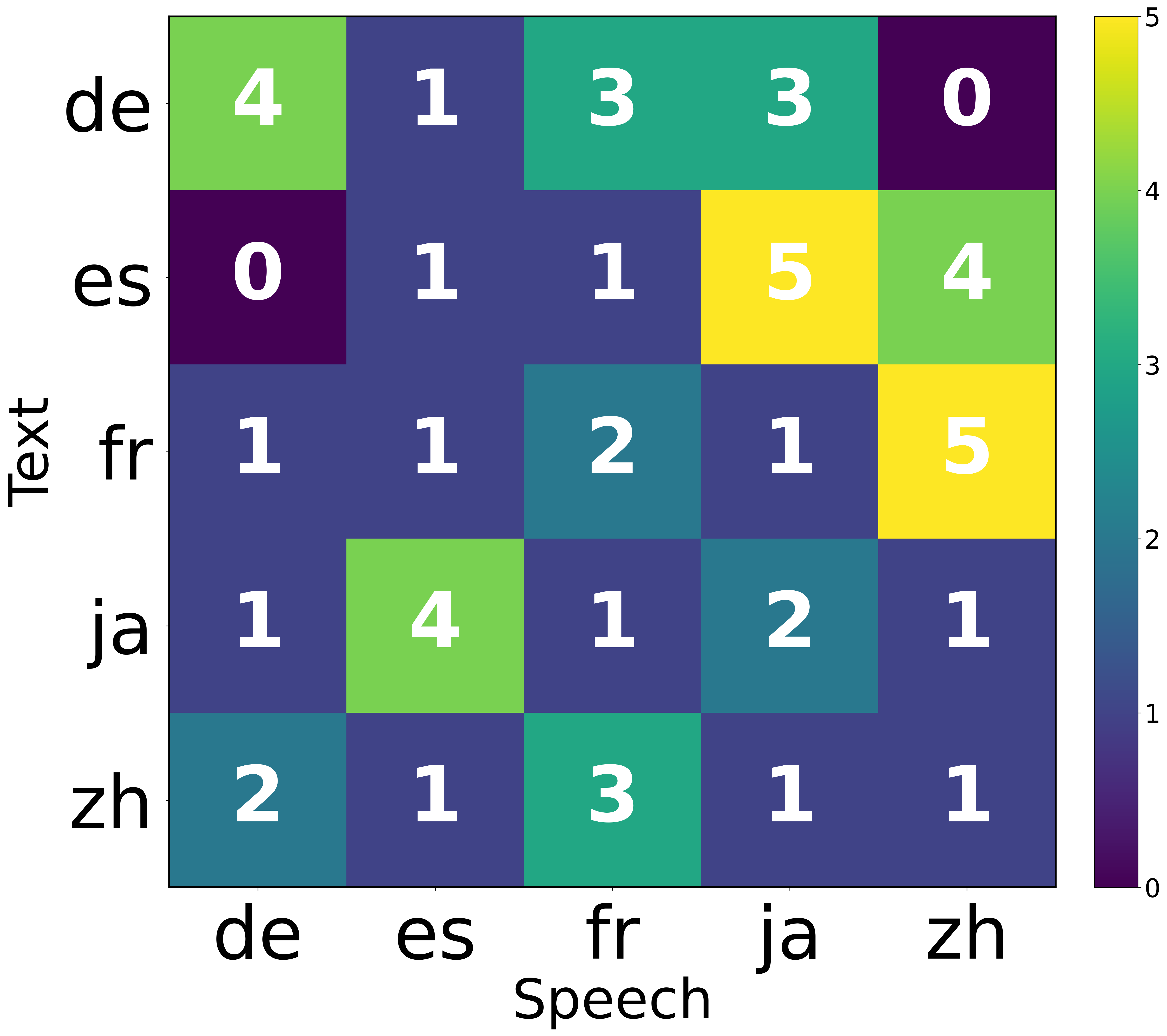
    }
    \caption{Mid, $t=11$}
\end{subfigure}
&
\begin{subfigure}[t]{0.23\linewidth}
    \centering
    \includegraphics[width=\linewidth]{
        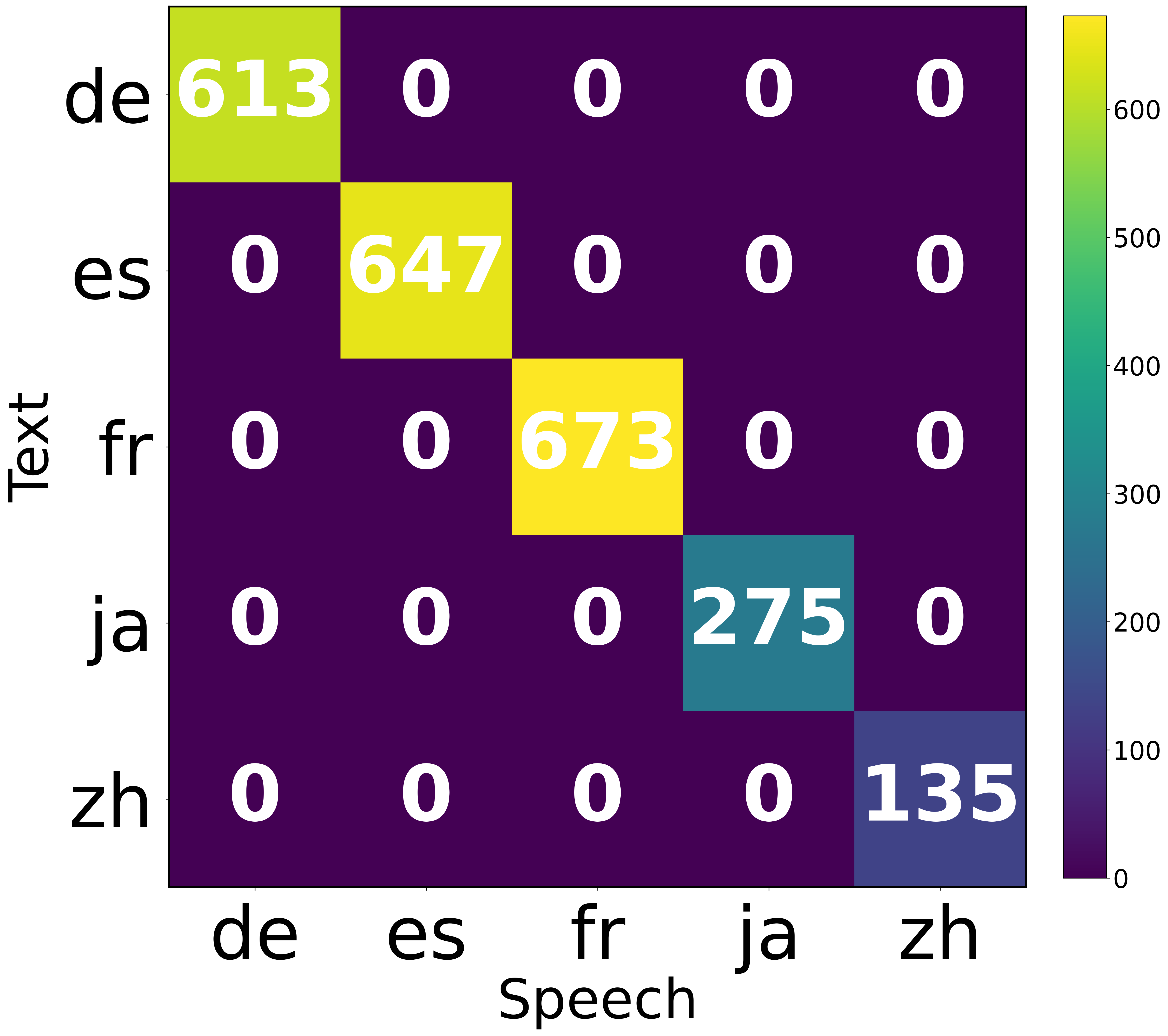
    }
    \caption{Top, $t=11$}
\end{subfigure}

\end{tabular}

\caption{
Cross-modal neuron overlap between speech and text in the decoder.
For SeamlessM4T, the strong same-language overlap at $t=1$ weakens
at later decoding steps, while mid-ranked neurons do not exhibit the
same diagonal structure. For comparison, Qwen2-Audio retains a strongly
diagonal overlap pattern at $t=11$.
}
\label{fig:overlap_heatmaps_textdec}
\end{figure*}

\section{Step-dependent cross-modal language Representations}
\label{sec:lang_rep}

We first present a detailed analysis of SeamlessM4T and then apply the same framework to the decoder-only model Qwen2-Audio as an additional case study.

\subsection{Cross-modal neuron overlap}
\label{subsec:neuron_overlap}

To investigate how language representations are shared across speech and text, we compare language-selective neurons identified independently from each modality at different decoding steps.

Table~\ref{tab:overlap_metrics} reveals substantial differences between the two models. SeamlessM4T shares only 5--7\% of language-selective neurons across modalities, whereas Qwen2-Audio consistently shares 45--47\%. The two models also exhibit markedly different overlap structures. In SeamlessM4T, same-language overlap is highest at the first decoding step (0.46) and decreases at later steps (0.18--0.21), while overlap between typologically related languages increases (from $-0.17$ to 0.19 and 0.16) without increasing overlap between unrelated languages ($-0.48$ to $-0.53$). In contrast, Qwen2-Audio maintains perfect same-language overlap (1.00) together with minimal typological overlap ($-0.47$) across all decoding steps, indicating stable language-specific cross-modal representations.

Figure~\ref{fig:overlap_heatmaps_textdec} illustrates these contrasting patterns. SeamlessM4T exhibits a strong same-language diagonal at the first decoding step that gradually weakens as overlap spreads toward typologically related languages. This structure is absent for random neurons sampled from the same layers and subcomponents, indicating that it is specific to language-selective neurons. By contrast, Qwen2-Audio remains almost perfectly diagonal throughout decoding, consistent with the quantitative metrics.

An exact permutation test confirmed significant same-language preference in both models. For SeamlessM4T, significance was observed at all decoding steps before correction (t = 1: p = 0.0083; t = 6, 11: p = 0.0417), but only the first decoding step remained significant after Bonferroni correction. In contrast, Qwen2-Audio showed significant same-language preference at every decoding step (all p = 0.0083), and all effects survived Bonferroni correction. No significant same-language preference is observed for the SeamlessM4T random-neuron baseline ($p=0.11$), whose apparent diagonal structure is driven by Chinese outliers (Appendix \ref{appendix:matrices}).

Overall, our analysis of SeamlessM4T reveals sparse but timestep-dependent neuron sharing. The additional analysis on Qwen2-Audio indicates that these behaviors are not universal across architectures.

\section{Cross-modal steering}
\label{sec:lang_control}

\subsection{Language control results}
\label{subsec:intervention_results}

\begin{table*}[t]
\centering
\small
\setlength{\tabcolsep}{3.5pt}
\begin{tabular}{lllrrrrrrrrr}
\toprule
\multirow{2}{*}{Task lang}
& \multirow{2}{*}{Task}
& \multirow{2}{*}{Intervention}
& \multicolumn{3}{c}{$t=1$}
& \multicolumn{3}{c}{$t=6$}
& \multicolumn{3}{c}{$t=11$} \\
\cmidrule(lr){4-6}
\cmidrule(lr){7-9}
\cmidrule(lr){10-12}
& & &
Random & Mid & Ours &
Random & Mid & Ours &
Random & Mid & Ours \\
\midrule

German & ASR & Japanese
& 0.00 & 0.00 & \textbf{0.00}
& 0.00 & 0.00 & \textbf{60.85}
& 0.00 & 0.00 & \textbf{83.32} \\

German & ASR & Chinese
& 0.00 & 0.00 & \textbf{0.00}
& 0.00 & 0.00 & \textbf{74.96}
& 0.00 & 0.00 & \textbf{85.81} \\

German & TR & Japanese
& 0.00 & 0.00 & \textbf{0.00}
& 0.00 & 0.00 & \textbf{14.16}
& 0.00 & 0.00 & \textbf{22.89} \\

German & TR & Chinese
& 0.00 & 0.00 & \textbf{0.00}
& 0.00 & 0.00 & \textbf{16.75}
& 0.00 & 0.00 & \textbf{31.67} \\

\midrule

French & ASR & Japanese
& 0.00 & 0.00 & \textbf{0.00}
& 0.00 & 0.00 & \textbf{65.34}
& 0.00 & 0.00 & \textbf{86.76} \\

French & ASR & Chinese
& 0.00 & 0.00 & \textbf{0.00}
& 0.00 & 0.00 & \textbf{77.73}
& 0.00 & 0.00 & \textbf{89.49} \\

French & TR & Japanese
& 0.00 & 0.00 & \textbf{0.00}
& 0.00 & 0.00 & \textbf{10.34}
& 0.00 & 0.00 & \textbf{25.96} \\

French & TR & Chinese
& 0.00 & 0.00 & \textbf{0.00}
& 0.00 & 0.00 & \textbf{30.70}
& 0.00 & 0.00 & \textbf{50.85} \\

\midrule

Japanese & ASR & German
& 0.10 & 0.10 & \textbf{0.10}
& 0.10 & 0.10 & \textbf{89.28}
& 0.10 & 0.10 & \textbf{95.43} \\

Japanese & ASR & French
& 0.10 & 0.10 & \textbf{0.10}
& 0.10 & 0.10 & \textbf{33.64}
& 0.10 & 0.10 & \textbf{43.78} \\

Japanese & TR & German
& 0.49 & 0.56 & \textbf{0.55}
& 0.49 & 0.49 & \textbf{63.12}
& 0.49 & 0.49 & \textbf{77.08} \\

Japanese & TR & French
& 0.65 & 0.56 & \textbf{0.59}
& 0.49 & 0.56 & \textbf{11.48}
& 0.49 & 0.38 & \textbf{24.03} \\

\midrule

Chinese & ASR & German
& 0.47 & 0.47 & \textbf{0.55}
& 0.55 & 0.55 & \textbf{91.69}
& 0.47 & 0.55 & \textbf{95.58} \\

Chinese & ASR & French
& 0.44 & 0.51 & \textbf{0.61}
& 0.47 & 0.69 & \textbf{51.43}
& 0.55 & 0.46 & \textbf{71.60} \\

Chinese & TR & German
& 1.89 & 1.89 & \textbf{1.89}
& 1.89 & 1.89 & \textbf{50.47}
& 1.79 & 1.89 & \textbf{58.96} \\

Chinese & TR & French
& 1.89 & 1.89 & \textbf{2.19}
& 1.89 & 1.83 & \textbf{15.09}
& 1.89 & 1.82 & \textbf{31.15} \\

\bottomrule
\end{tabular}

\caption{
Target-script rate (\%) under random-neuron, mid-ranked-neuron, and
language-control-neuron interventions.
For German and French task outputs, the table reports the CJK-script rate
(Latin-to-CJK control); for Japanese and Chinese task outputs, it reports the
Latin-script rate (CJK-to-Latin control).
\textit{Random} denotes intervention on randomly selected neurons,
\textit{Mid} denotes intervention on mid-ranked neurons, and
\textit{Ours} denotes intervention on the identified language-control neurons.
Values for our method are highlighted in bold.
}
\label{tab:en2x_unimodal_neuron_scripts}
\end{table*}


To evaluate whether language-control neurons causally influence output language,
we examine script changes under intervention across decoding steps. We present the SeamlessM4T results in the main paper and the corresponding Qwen2-Audio results in Appendix~\ref{appendix:language-control-qwen}.

Table~\ref{tab:en2x_unimodal_neuron_scripts} shows that effective language control emerges only at later decoding steps. At $t=1$, interventions produce little control. In contrast, substantial script shifts appear at $t=6$ and become even stronger at $t=11$. For example, Chinese ASR controlled toward German reaches 95.58\% Latin script at $t=11$, and these effects transfer to text repetition.

Consistently, the mean cross-modal effect ratio increases from 0.36 at $t=6$ to 0.49 at $t=11$. Both the source-modality (ASR) and target-modality (text repetition) control rates increase over this interval (68.12\%$\rightarrow$81.47\% and 26.51\%$\rightarrow$40.32\%, respectively), with a proportionally larger improvement in text repetition. This indicates that later-step language-control neurons not only become more effective in the source modality but also transfer more effectively across modalities.


\subsection{Qualitative evaluation}

The script-based metric only measures whether the output shifts to the target writing system and therefore cannot distinguish genuine language transfer from superficial script changes. We therefore perform a qualitative evaluation of intervention outputs.

Table~\ref{tab:de-zh-intervention-example} shows a representative German$\rightarrow$Chinese intervention. The intervention successfully shifts the output toward the target language while largely preserving semantic content. Additional examples are provided in Appendix~\ref{appendix:qualitative}.

To evaluate this behavior systematically, we manually inspected 100 outputs from German$\leftrightarrow$Chinese interventions generated using our language-control neurons ($t=1,6,11$), together with random and mid-neuron baselines ($t=11$). Each token was annotated as (i) source-language and semantically related to the reference (Src Rel), (ii) target-language and semantically related (Tgt Rel), (iii) source-language but semantically unrelated (Src Unrel), (iv) target-language but semantically unrelated (Tgt Unrel), or (v) another language (Other Lang). We additionally annotated sentence-level semantic coherence and repetitive collapse.

As shown in Table~\ref{tab:specificity_baselines}, unrelated tokens were rare (0--5\%) and semantic coherence remained high (75--100\%) after excluding repetitive-collapse cases. Tokens from other languages were also infrequent, indicating that the intervention selectively increases target-language content rather than merely producing the correct script while largely preserving semantic meaning.

\begin{table*}[t]
\centering
\small
\resizebox{\linewidth}{!}{%
\begin{tabular}{llrrrrrrrrr}
\toprule
Condition & Task & $t$ &
Src-Rel & Tgt-Rel &
Src-Unrel & Tgt-Unrel &
Other Lang &
Semantic Coherence &
Without Collapse \\
\midrule
OURS   & S2T & 1  & 95.95 & \textbf{0.00} & 4.05 & 0.00 & 0.00 & 88.89 & 90.00 \\
OURS   & S2T & 6  & 11.65 & \textbf{81.55} & 0.00 & 0.00 & 6.80 & 80.00 & 50.00 \\
OURS   & S2T & 11 & 8.82 & \textbf{84.31} & 0.00 & 3.92 & 2.94 & 80.00 & 50.00 \\
OURS   & T2T & 1  & 97.20 & \textbf{2.80} & 0.00 & 0.00 & 0.00 & 100.00 & 100.00 \\
OURS   & T2T & 6  & 68.45 & \textbf{30.48} & 0.00 & 1.07 & 0.00 & 85.71 & 70.00 \\
OURS   & T2T & 11 & 54.27 & \textbf{42.71} & 0.00 & 2.01 & 1.01 & 75.00 & 80.00 \\
\midrule
MID    & S2T & 11 & 95.59 & \textbf{0.00} & 4.41 & 0.00 & 0.00 & 88.89 & 90.00 \\
MID    & T2T & 11 & 97.20 & \textbf{2.80} & 0.00 & 0.00 & 0.00 & 100.00 & 100.00 \\
\midrule
RANDOM & S2T & 11 & 95.98 & \textbf{0.00} & 4.02 & 0.00 & 0.00 & 88.89 & 90.00 \\
RANDOM & T2T & 11 & 97.20 & \textbf{2.80} & 0.00 & 0.00 & 0.00 & 100.00 & 100.00 \\
\bottomrule
\end{tabular}
}
\caption{Qualitative evaluation of German$\leftrightarrow$Chinese interventions. We manually annotated 100 outputs in total. ``Src Rel'' and ``Tgt Rel'' denote source-language and target-language tokens that are semantically consistent with the reference, respectively. ``Src Unrel'' and ``Tgt Unrel'' denote source-language and target-language tokens that are semantically unrelated to the reference. ``Other Lang'' denotes tokens from languages other than the source or target. ``Semantic Coherence'' reports the percentage of outputs preserving the original meaning, excluding outputs with repetitive collapse. ``Without Collapse'' reports the percentage of outputs that avoid degenerate repetitive generation.}
\label{tab:specificity_baselines}
\end{table*}

\begin{table}[t]
\centering
\small
\setlength{\tabcolsep}{4pt}
\begin{tabular}{p{0.16\linewidth}p{0.78\linewidth}}
\toprule
\textbf{Condition} & \textbf{Output} \\
\midrule
\textbf{Gold} & in entwickelten ländern werden sie selten ähnlich viele beschwerden über die wasserqualität oder den einsturz von brücken hören \\
\textbf{ASR} & \zh{在发达的国家,你会你很少类似许多投诉关于水质或桥梁的崩塌的听} \\
\textbf{Text repetition} & in entwickelten Ländern werden sie\zh{很少类似地许多} Beschwerden über die\zh{水质} oder den einsturz von brücken \\
\bottomrule
\end{tabular}
\caption{Example of German-to-Chinese intervention with neurons identified from ASR. The intervention shifts the output toward Chinese in ASR and induces mixed-language output in text repetition.}
\label{tab:de-zh-intervention-example}
\end{table}

\section{Discussion}

\paragraph{Input-side: model-dependent cross-modal processing.}

Our results reveal fundamentally different forms of cross-modal language processing. SeamlessM4T shares only 5--7\% of language-selective neurons between speech and text, whereas Qwen2-Audio shares nearly half, indicating substantially different degrees of cross-modal integration.

One possible explanation is the different alignment between speech and text representations before decoding. Qwen2-Audio initializes its speech encoder from Whisper \citep{radford2022robustspeechrecognitionlargescale,chu2024qwen2audiotechnicalreport}, whose speech-to-text pretraining may produce representations that are already well aligned with text. In contrast, SeamlessM4T relies on a speech-only pretrained speech encoder and exhibits timestep-dependent reorganization of language representations, whereas Qwen2-Audio maintains stable organization throughout decoding. Although speculative, these findings suggest that both the extent of cross-modal integration and its temporal evolution depend strongly on model architecture and pretraining.

\paragraph{Output-side: partially shared control mechanisms.}
In contrast, language-control neurons exhibit a different pattern. Steering using neurons identified at later decoding steps ($t=6$, $t=11$) is substantially more effective than steering using neurons identified at $t=1$ and transfers from ASR to text generation. This provides causal evidence that output-language control relies on partially shared mechanisms across modalities.

Moreover, the cross-modal effect ratio increases at later decoding steps. Both the source-modality (ASR) and target-modality (text repetition) control rates increase from $t=6$ to $t=11$, with a proportionally larger improvement in text repetition, indicating that later-step language-control neurons transfer more effectively across modalities. One possible explanation is that, as autoregressive context accumulates, language-control signals become increasingly independent of the input modality.




\section{Conclusion}

We introduced a generation-step-aware framework for analyzing cross-modal language processing in multilingual speech--text models by separating language-representation and language-control neurons.

Applying this framework to SeamlessM4T, we showed that language representations undergo generation-step-dependent reorganization, whereas language-control neurons identified at later decoding steps transfer causally across modalities, providing evidence for partially shared mechanisms for output-language control. Extending the analysis to the decoder-only model Qwen2-Audio further showed that these patterns do not necessarily generalize across architectures. Together, our findings demonstrate the importance of generation-step-aware analyses for understanding multilingual speech--text models.

\section{Limitations}
Our study has several limitations.

First, our empirical evaluation is limited to two multilingual speech--text models: SeamlessM4T v2 and Qwen2-Audio. While these models differ substantially in architecture and training, they cannot fully represent the broad design space of multilingual multimodal systems. Extending our analysis to additional architectures would help determine the generality of the generation-step-dependent behaviors observed in this work.

Second, our framework relies on AP-based neuron identification, which captures statistical associations between neuron activations and linguistic labels but does not establish causal necessity. Although we complement this with intervention experiments, our conclusions remain constrained by this framework.

Third, we use synthetic speech to control for acoustic variability. While this improves comparability, future work should verify whether similar patterns hold for natural speech.

Finally, our evaluation setup introduces asymmetries across tasks. In particular, language tags in ASR and text repetition might introduce some noise in the neuron identification step, and ASR and text repetition differ in their linguistic objectives, which may affect neuron identification and cross-modal comparisons. More symmetric settings, such as $En \rightarrow X$ translation across modalities, should be explored in future work.

\section*{Acknowledgments}
We thank anonymous reviewers for helpful feedbacks and discussions.
We thank Yifan Wang and Leonie Weissweiler for helpful feedbacks during revisions.


\section*{Ethical considerations}
This work evaluates internal processing mechanisms of multilingual speech--text models via activation recording and neuron-level interventions. It does not involve human subjects, user interaction, or deployment in real-world decision-making systems.

We use publicly available resources, namely SeamlessM4T v2, Qwen2-Audio and the FLEURS dataset, in accordance with their respective licenses. Speech inputs are generated via text-to-speech synthesis using XTTS v2 conditioned on a single speaker. The synthesized audio does not contain real personal data, and we do not reveal or infer any identifying information about individuals.

\bibliography{custom}

\appendix
\section{Appendix}
We present additional details and results which complement our main text.


\subsection{Synthesized audio quality}
\label{appdendix:audioquality}

As described in Subsection~\ref{subsec:data_preprocessing}, we re-synthesized the FLEURS utterances using a text-to-speech (TTS) model. For all six languages (English, German, Spanish, French, Japanese, and Mandarin), we manually inspected the generated audio. For five languages, the speech was natural and fully intelligible. For Mandarin, the audio was generally understandable but occasionally exhibited pronunciation errors.

\paragraph{Illustrative example.}
The following Mandarin sentence from FLEURS is shown in pinyin only (Chinese characters omitted for clarity):

\begin{tcolorbox}[
  colback=gray!5,
  colframe=gray!50,
  boxrule=0.5pt,
  arc=3pt,
  left=6pt,
  right=6pt,
  top=6pt,
  bottom=6pt
]
\textbf{Target sentence (pinyin):}\\
Hàn Mì’ěrdùn quèrèn Huò Huá Dé dàxué yīyuàn shōuzhì de bìngrén qíngkuàng wěndìng.\\[0.5em]

\textbf{Gloss:}\\
Hamilton confirm Howard University hospital admit-\textsc{de} patient condition stable.\\[0.5em]

\textbf{Intended meaning:}\\
``Hamilton confirmed that the condition of the patients admitted by Howard University Hospital was stable.''
\end{tcolorbox}

\noindent\textbf{Observed synthesis errors.}
The synthesized speech deviates from the target pronunciation in several ways.
\textbf{Phoneme substitution:}
\emph{quèrèn} $\rightarrow$ \emph{kuèrèn}, introducing an illicit onset in Mandarin.
\textbf{Word-boundary errors:}
\emph{bìngrén qíngkuàng} (“patient condition”) $\rightarrow$ \emph{bìngrénqíng kuàng}.
\textbf{Syllable insertion:}
\emph{de} $\rightarrow$ \emph{dede}, \emph{wěndìng} $\rightarrow$ \emph{wěndìnggua}.

\paragraph{Impact on evaluation.}
Although these errors indicate limitations in synthetic Mandarin speech, they do not affect sentence-level intelligibility and occur only sporadically. Since our study focuses on cross-lingual and cross-modal representational patterns rather than fine-grained phonetic accuracy, we expect their impact on the reported results to be minimal.

\subsection{Formulation for decoder-only models}
\label{appdendix:formulation}

For decoder-only models, there is no encoder component. Let
$X_{\mathrm{dec},t}=(y_1,\ldots,y_{t-1})$
denote the decoder input available when predicting the $t$-th output token. For layer $l$ and submodule $m$, let
$\mathbf{A}_{\mathrm{dec}}^{(l,m)}(X_{\mathrm{dec},t})\in\mathbb{R}^{t\times d}$
denote the decoder activations.

Following the same formulation as in the encoder--decoder setting, neuron activity at generation step $t$ is defined as the averaged activation at the current decoding position:

\[
\bigl[
\mathbf{s}_{\mathrm{dec}}^{(l,m)}
\bigr]_j
=
\frac{1}{t}
\sum_{i=1}^{t}
\bigl[
\mathbf{A}_{\mathrm{dec}}^{(l,m)}(X_{\mathrm{dec},t})
\bigr]_{i,j}.
\]

Average-precision (AP) scores are then computed identically to the encoder--decoder formulation:

\[
\mathrm{AP}^{(l,m)}_j
=
\mathrm{AP}(\mathbf{s}^{(l,m)}_j,\mathbf{z}),
\]

where $\mathbf{z}$ denotes the binary language labels. Neurons are ranked separately for each decoding step using the resulting AP scores.

\subsection{Distribution of language neurons}
\label{appendix:dist_lang_neuron}
Having established the cross-modal evaluation framework, we next examine how language-representation neurons are distributed in SeamlessM4T and Qwen2-Audio. For each module and decoder step, we report the distributions of top-1{,}000 neurons identified from German speech and German text, respectively.

\begin{figure*}[t]
\centering

\begin{subfigure}{0.48\linewidth}
\centering
\includegraphics[width=\linewidth]{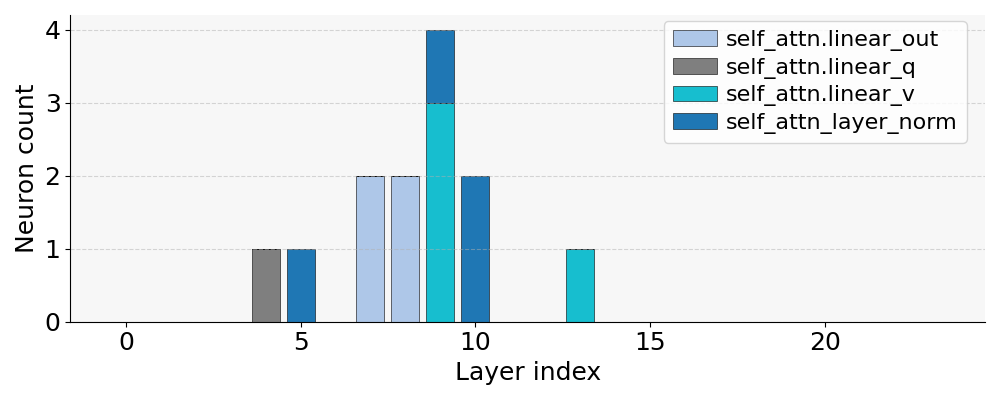}
\caption{Speech encoder (Attention)}
\end{subfigure}\hfill
\begin{subfigure}{0.48\linewidth}
\centering
\includegraphics[width=\linewidth]{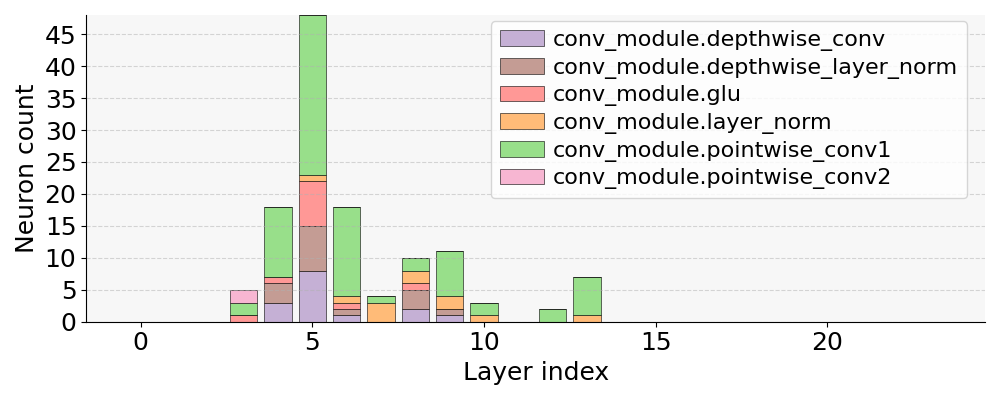}
\caption{Speech encoder (Convolution)}
\end{subfigure}


\begin{subfigure}{0.48\linewidth}
\centering
\includegraphics[width=\linewidth]{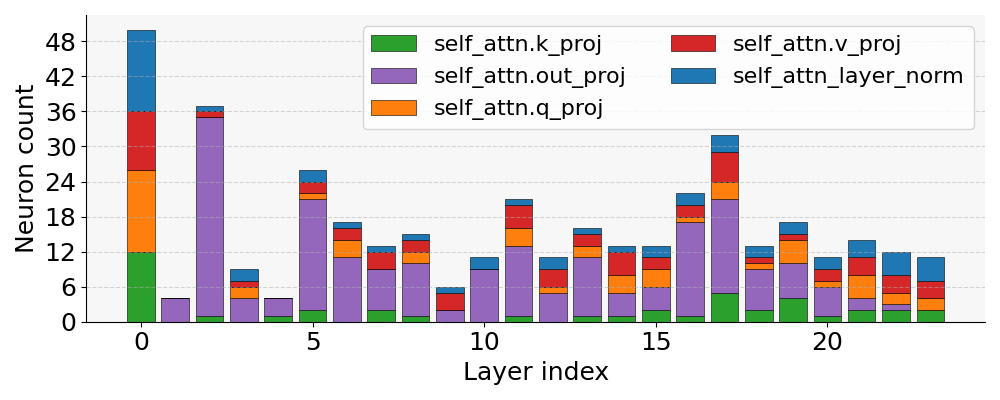}
\caption{Text encoder (Attention)}
\end{subfigure}\hfill
\begin{subfigure}{0.48\linewidth}
\centering
\includegraphics[width=\linewidth]{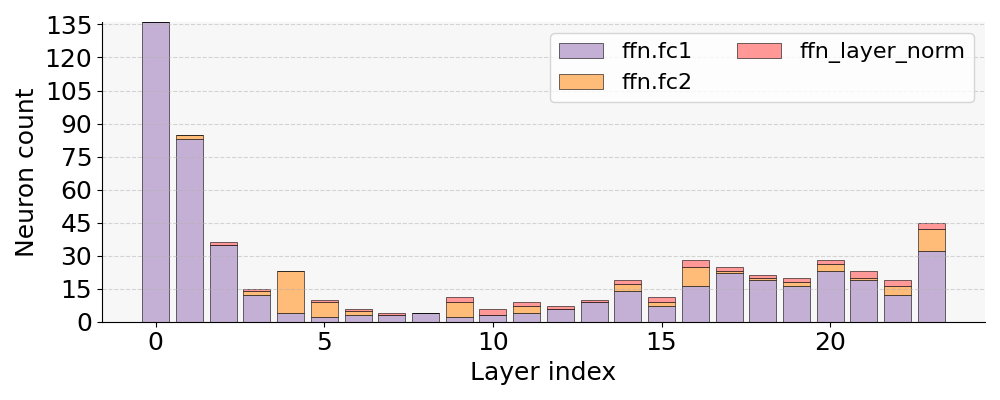}
\caption{Text encoder (FFN)}
\end{subfigure}

\caption{Layer-wise distribution of top-1{,}000 language-representation neurons identified from SeamlessM4T. In the speech encoder, neurons concentrate in middle layers, whereas in the text encoder they peak at the first layer, reflecting that language information must be extracted from continuous acoustic input in speech but is already explicit in discrete text tokens.}
\label{fig:encoder_distribution}
\end{figure*}

\begin{figure*}[t]
\centering

\begin{subfigure}{0.48\linewidth}
\centering
\includegraphics[width=\linewidth]{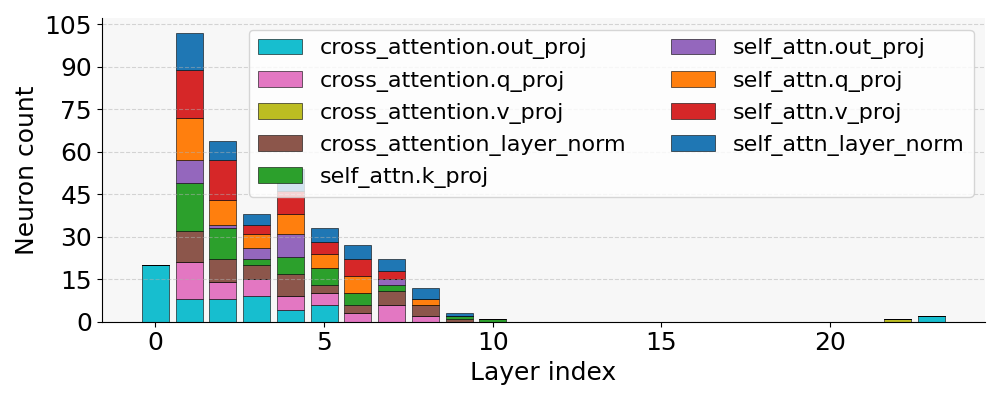}
\caption{Speech $t=1$ Attention}
\end{subfigure}\hfill
\begin{subfigure}{0.48\linewidth}
\centering
\includegraphics[width=\linewidth]{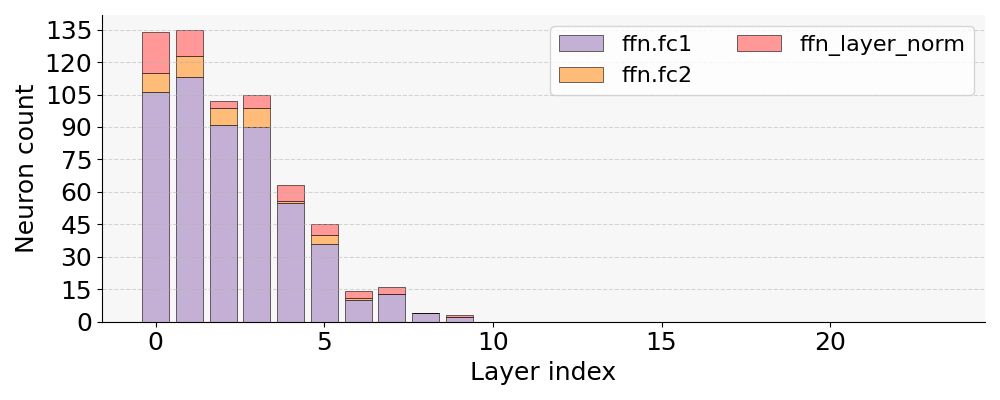}
\caption{Speech $t=1$ FFN}
\end{subfigure}


\begin{subfigure}{0.48\linewidth}
\centering
\includegraphics[width=\linewidth]{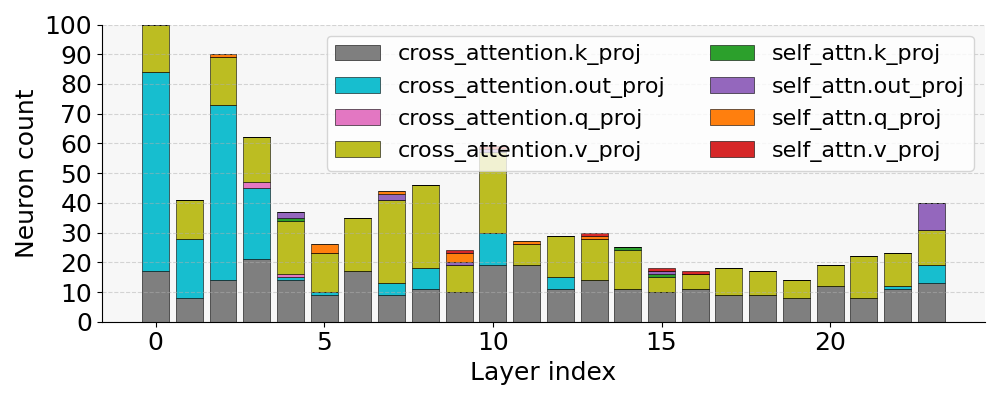}
\caption{Speech $t=11$ Attention}
\end{subfigure}\hfill
\begin{subfigure}{0.48\linewidth}
\centering
\includegraphics[width=\linewidth]{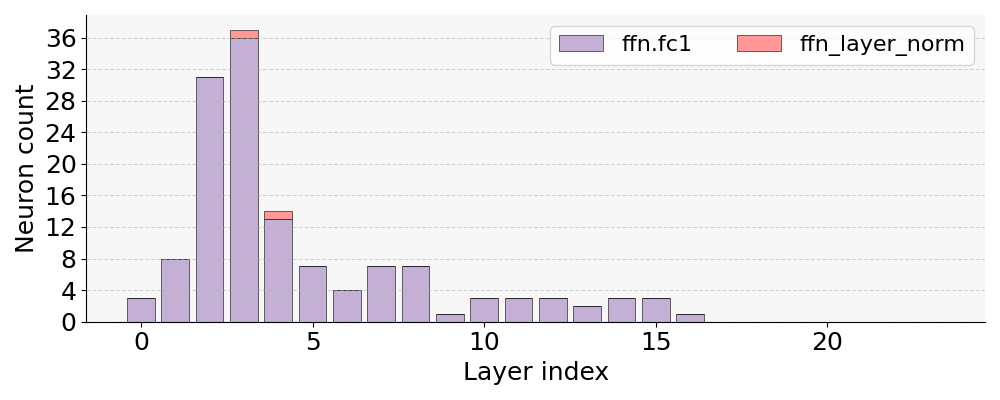}
\caption{Speech $t=11$ FFN}
\end{subfigure}

\caption{Distribution of top-1{,}000 language-representation neurons in the SeamlessM4T decoder across decoding steps (from German speech). Language selectivity shifts from self-attention at $t=1$ to cross-attention at $t=11$, indicating a transition from encoder-driven to autoregressive processing.}
\label{fig:decoder_distribution}
\end{figure*}

\begin{figure*}[t]
\centering

\begin{subfigure}{0.48\linewidth}
\centering
\includegraphics[width=\linewidth]{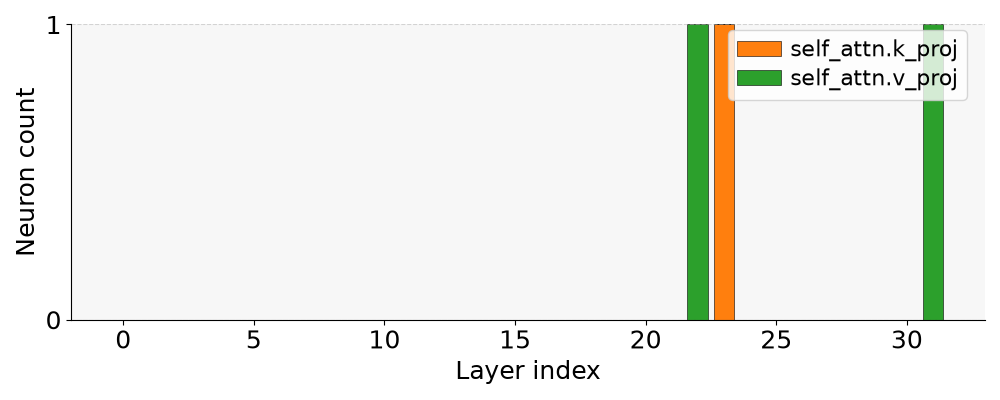}
\caption{Speech $t=1$ Attention}
\end{subfigure}\hfill
\begin{subfigure}{0.48\linewidth}
\centering
\includegraphics[width=\linewidth]{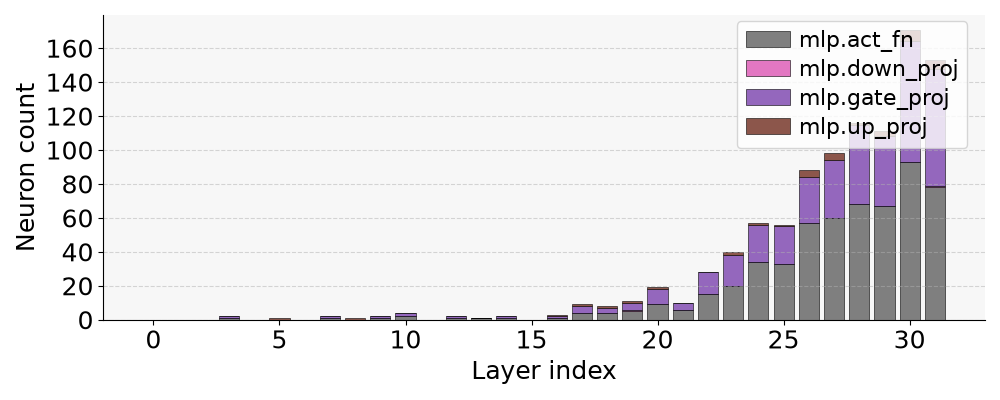}
\caption{Speech $t=1$ FFN}
\end{subfigure}


\begin{subfigure}{0.48\linewidth}
\centering
\includegraphics[width=\linewidth]{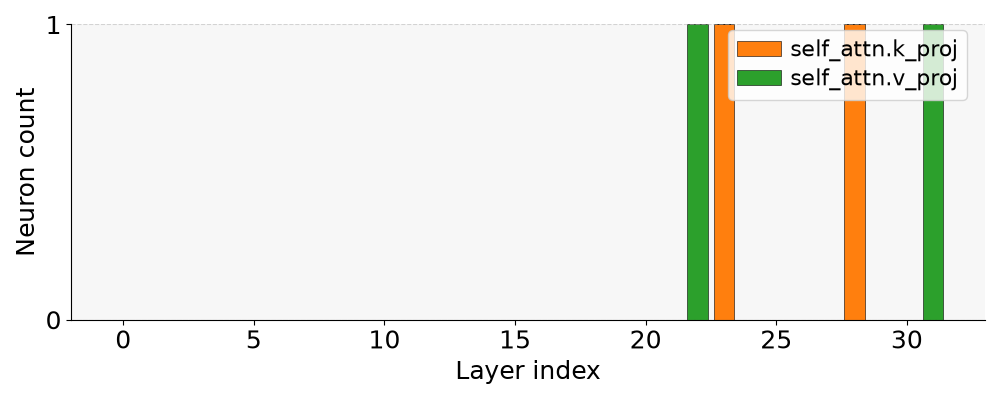}
\caption{Speech $t=11$ Attention}
\end{subfigure}\hfill
\begin{subfigure}{0.48\linewidth}
\centering
\includegraphics[width=\linewidth]{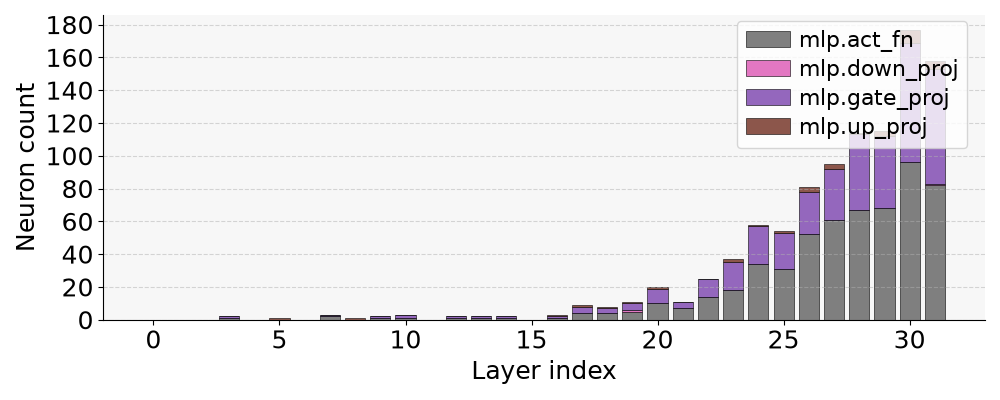}
\caption{Speech $t=11$ FFN}
\end{subfigure}

\caption{Distribution of top-1,000 language-representation neurons in the Qwen2-Audio decoder across decoding steps (from German speech). Language-selective neurons remain stably concentrated in the late FFN layers, with little change between $t=1$ and $t=11$.}
\label{fig:qwen_decoder_distribution}
\end{figure*}

\paragraph{SeamlessM4T Speech encoder.}
Figures~\ref{fig:encoder_distribution}(a) and (b) show that language-representation neurons concentrate in middle layers (approximately 4--10), indicating that language-dependent processing emerges after initial acoustic processing and becomes more abstract in later layers.

We also observe component-level skew, where certain pointwise convolutional submodules contain more language-selective neurons than expected, 
while attention sublayers contribute comparatively fewer.

\paragraph{SeamlessM4T Text encoder.}
Figures~\ref{fig:encoder_distribution}(c) and (d) show that language-representation neurons peak in the first layer, 
indicating that language identity is primarily captured by surface-form cues such as script and lexical statistics. Unlike speech, text does not involve speaker or prosodic variation, making early-layer features highly diagnostic of language identity.

\paragraph{SeamlessM4T Text decoder.}
Figures~\ref{fig:decoder_distribution} show that language selectivity shifts from self-attention at $t=1$ (a) to cross-attention at $t=11$ (c), indicating a transition from encoder-driven processing to autoregressive, context-dependent computation.

At \(t=1\), language selectivity is concentrated primarily in self-attention (query, key, value, and layer normalization), whereas at \(t=11\), it shifts predominantly to cross-attention (key, value, and output projections).
These results indicate that the focus of language processing in the decoder is strongly position-dependent, rather than fixed across generation.

\paragraph{Qwen2-Audio decoder.}
In contrast to SeamlessM4T, Figures~\ref{fig:qwen_decoder_distribution} show that Qwen2-Audio exhibits largely time-invariant neuron distributions across decoding steps. Language-selective neurons remain consistently concentrated in the later FFN layers rather than in attention modules, indicating little functional reorganization during generation.

\subsection{Spearman correlation evaluation}
\label{appendix:spearman}

\begin{figure*}[t]
\centering

\begin{subfigure}{0.32\linewidth}
\includegraphics[width=\linewidth]{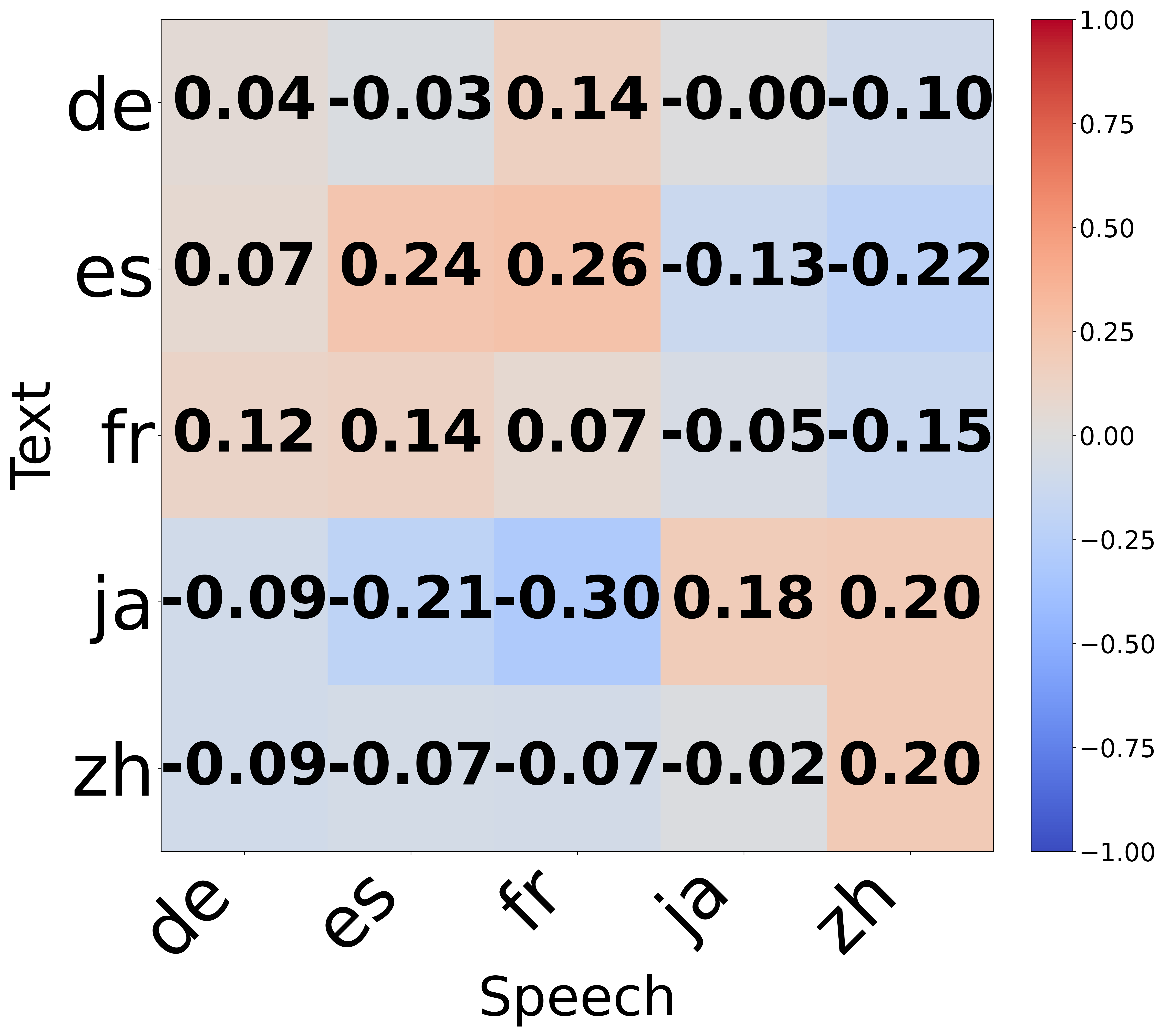}
\caption{SeamlessM4T, $t=1$}
\end{subfigure}\hfill
\begin{subfigure}{0.32\linewidth}
\includegraphics[width=\linewidth]{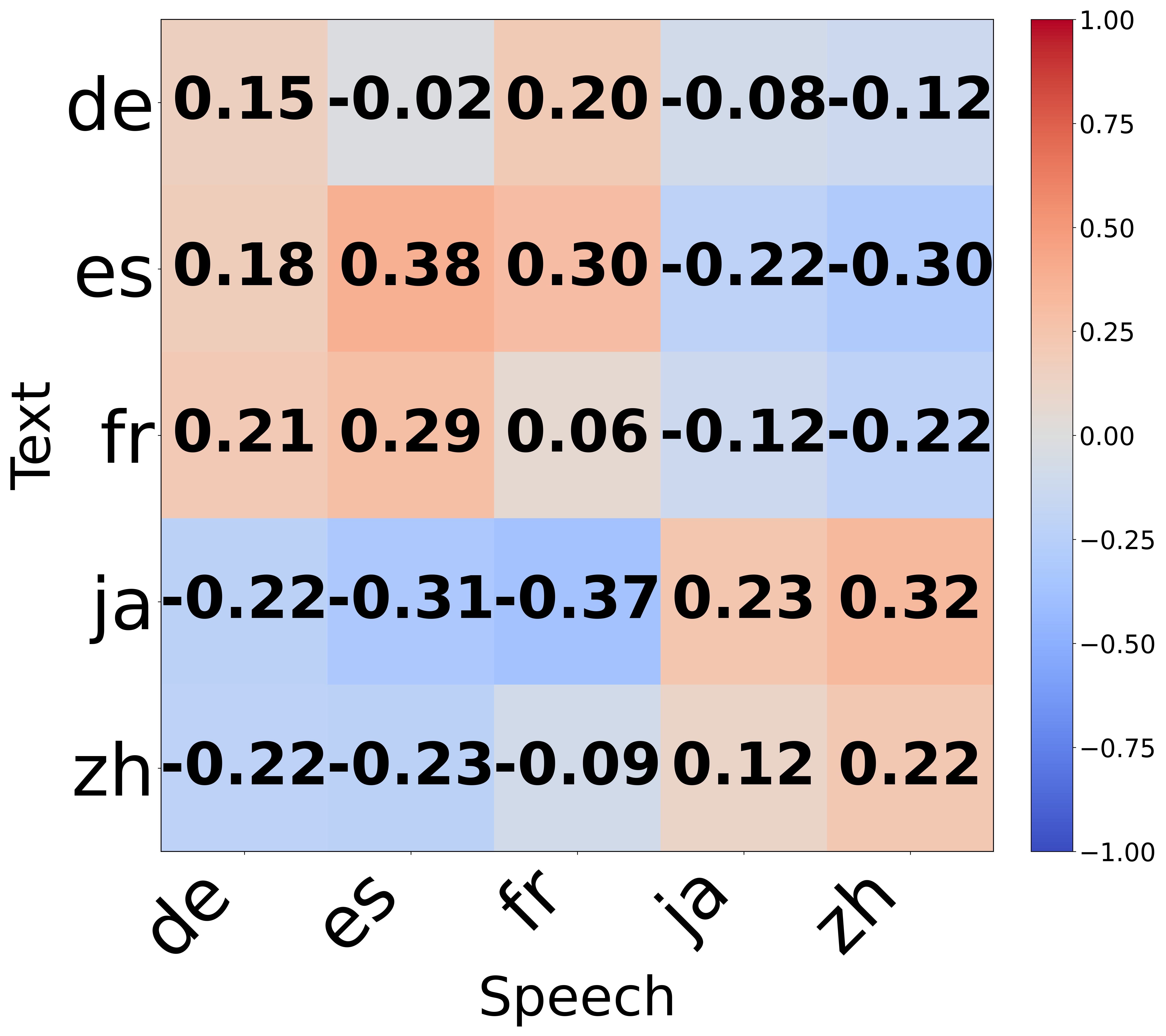}
\caption{SeamlessM4T, $t=6$}
\end{subfigure}\hfill
\begin{subfigure}{0.32\linewidth}
\includegraphics[width=\linewidth]{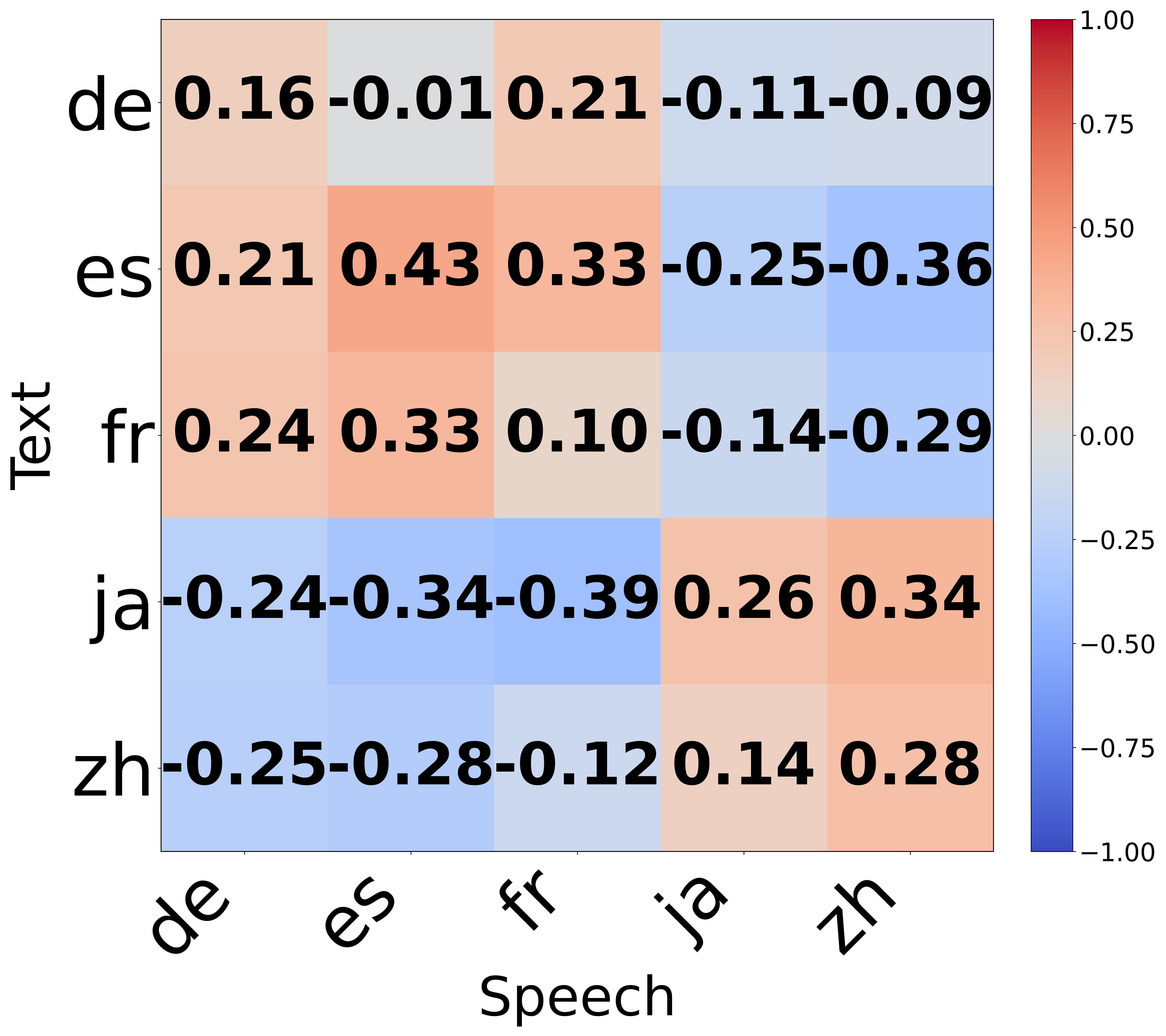}
\caption{SeamlessM4T, $t=11$}
\end{subfigure}

\begin{subfigure}{0.32\linewidth}
\includegraphics[width=\linewidth]{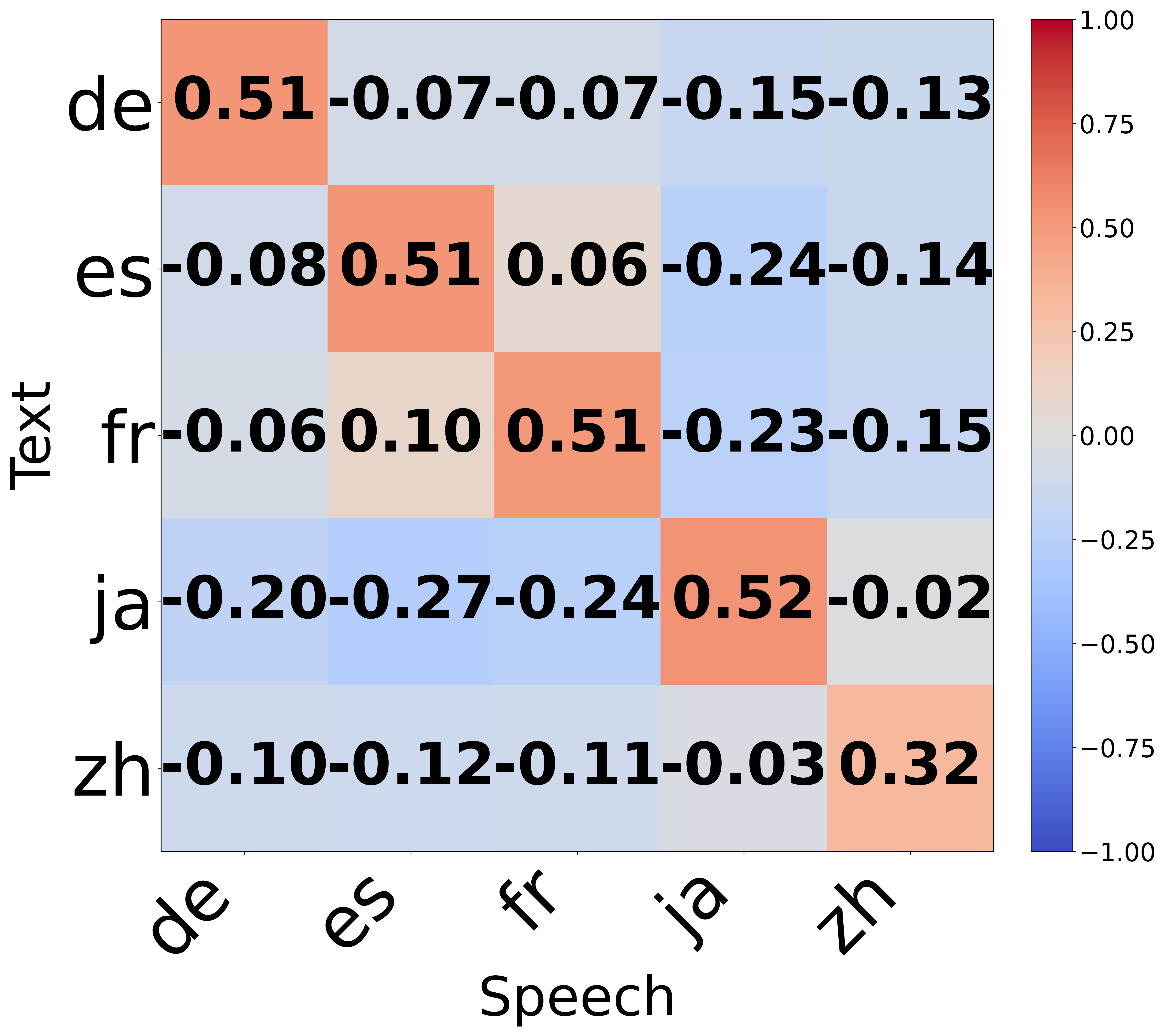}
\caption{Qwen2-Audio, $t=1$}
\end{subfigure}\hfill
\begin{subfigure}{0.32\linewidth}
\includegraphics[width=\linewidth]{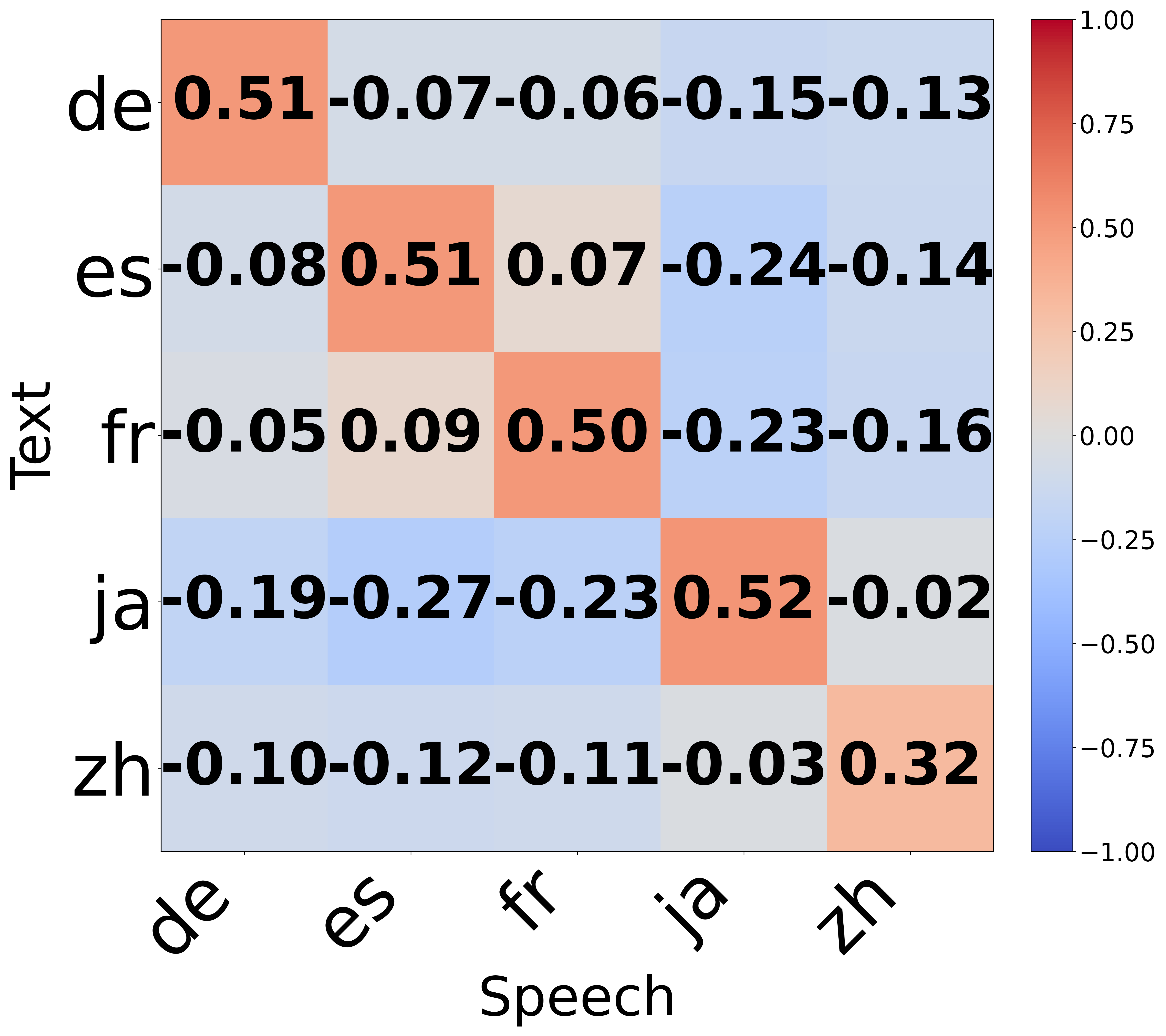}
\caption{Qwen2-Audio, $t=6$}
\end{subfigure}\hfill
\begin{subfigure}{0.32\linewidth}
\includegraphics[width=\linewidth]{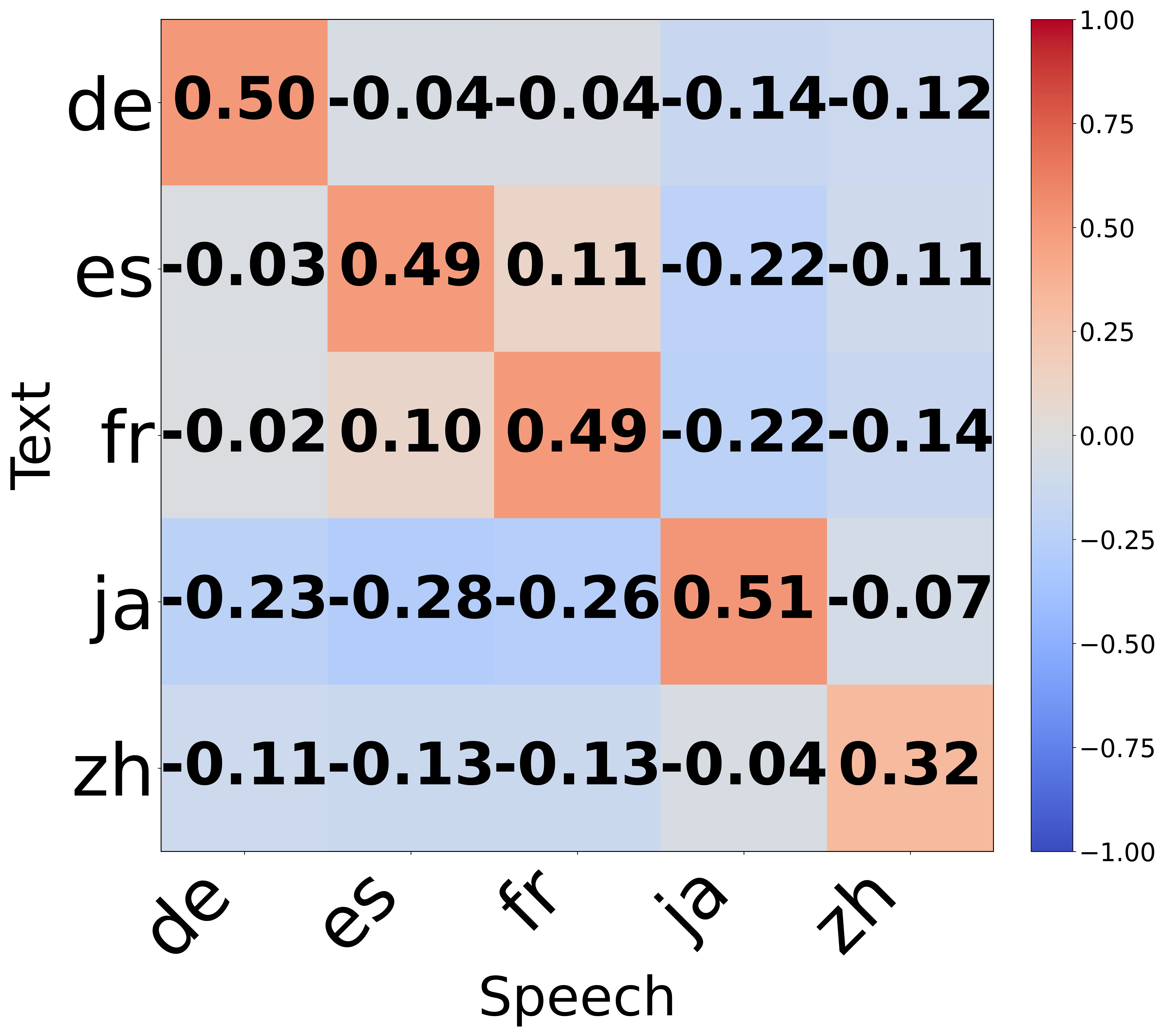}
\caption{Qwen2-Audio, $t=11$}
\end{subfigure}

\caption{
Spearman correlations between speech- and text-conditioned neuron activations
in the decoder of SeamlessM4T (top) and Qwen2-Audio (bottom).
SeamlessM4T exhibits progressively stronger typological organization at later
decoding steps, whereas Qwen2-Audio maintains stable same-language correlations
throughout decoding.
}
\label{fig:spearman_heatmaps_textdec}
\end{figure*}

To assess global cross-modal similarity beyond top-ranked neurons, we compute Spearman correlations over the full set of neuron activations for SeamlessM4T and Qwen2-Audio.

Figure~\ref{fig:spearman_heatmaps_textdec} shows that the global correlation patterns are broadly consistent with the overlap-based analysis. In SeamlessM4T, typological structure becomes more pronounced as decoding progresses. At later decoding steps, some language pairs within the same typological group exhibit stronger cross-modal correlations than the corresponding same-language pairs, suggesting that global activation patterns are increasingly organized according to language relatedness.

In contrast, Qwen2-Audio consistently exhibits the strongest correlations for same-language speech--text pairs across decoding steps. Although weak typological structure is visible for a small number of language pairs, such as Spanish and French, these effects remain comparatively limited.

Overall, these results suggest that the global organization of cross-modal language representations is also model-dependent.

\subsection{Sensitivity to $k$ in overlap evaluation}

\begin{figure*}[t]
\centering

\begin{subfigure}[t]{0.49\linewidth}
    \centering
    \includegraphics[width=\linewidth]{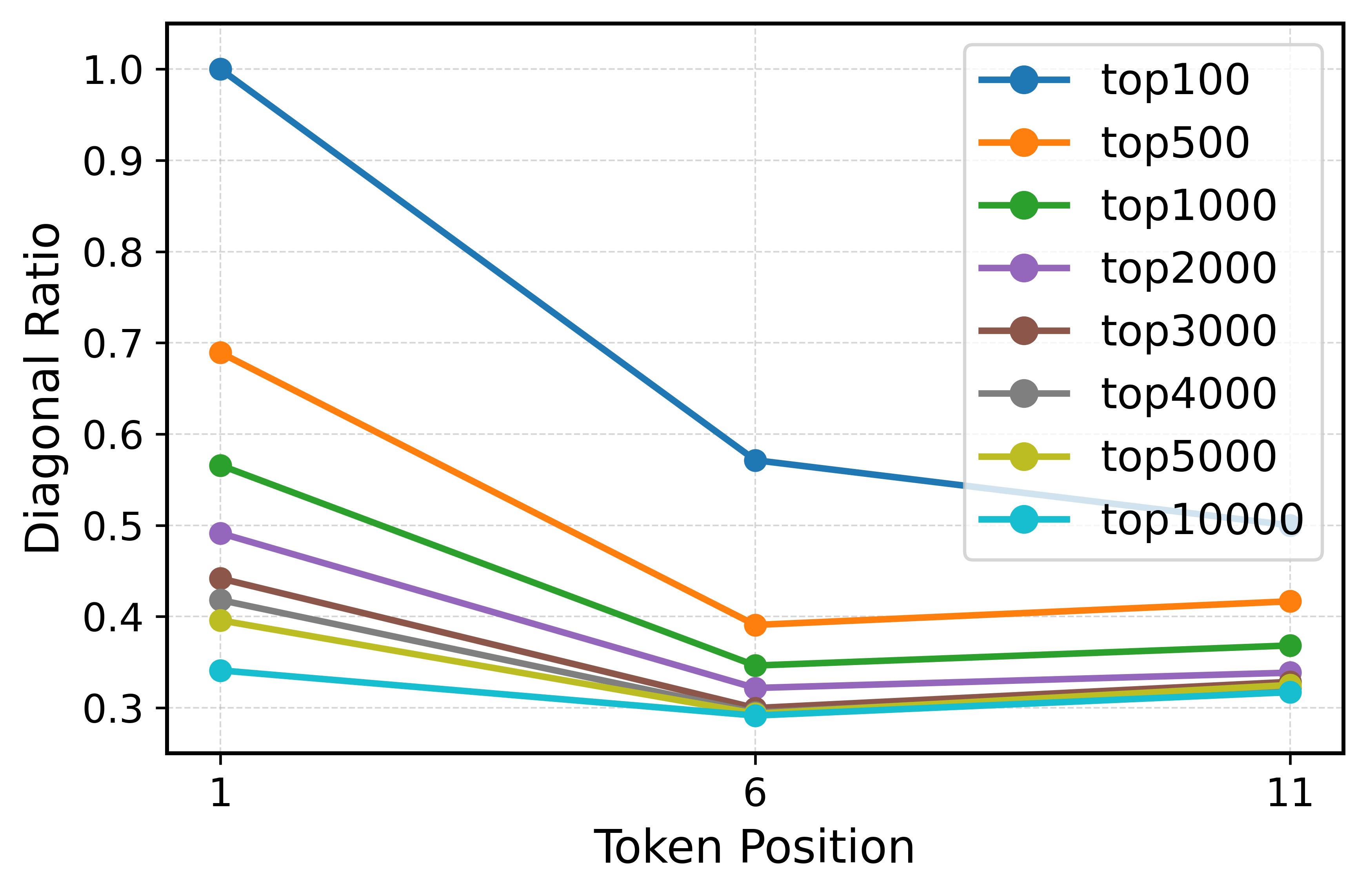}
    \caption{SeamlessM4T}
\end{subfigure}\hfill
\begin{subfigure}[t]{0.49\linewidth}
    \centering
    \includegraphics[width=\linewidth]{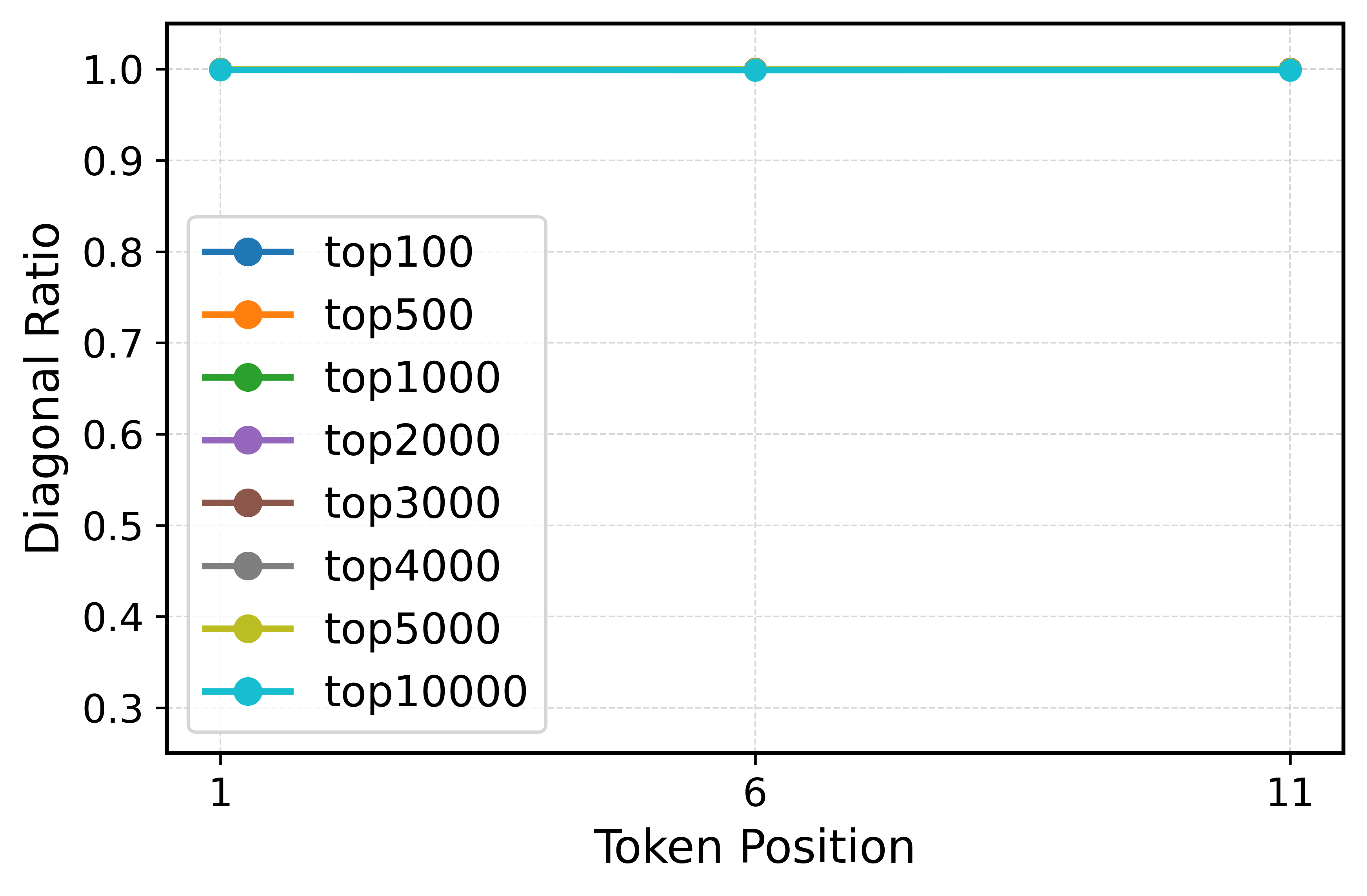}
    \caption{Qwen2-Audio}
\end{subfigure}

\caption{
Sensitivity of cross-modal overlap to the choice of $k$.
For each model, we report the diagonal ratio of the overlap matrix (i.e., the proportion of same-language overlap) across decoding steps.
In SeamlessM4T (left), the diagonal ratio is consistently highest at $t=1$, decreases at $t=6$, and partially recovers at $t=11$ (except for $k=100$), indicating that early decoding relies on a relatively small subset of strongly aligned cross-modal neurons.
In contrast, Qwen2-Audio (right) exhibits a different trend, showing that the effect of $k$ on cross-modal alignment is substantially more stable across decoding steps.
}
\label{fig:diagonal_ratio_plot}
\end{figure*}

To assess the robustness of our results to the choice of \(k\), we analyze the raw diagonal ratio of the cross-modal overlap matrix (before chance correction) across different values of \(k\).

Figure~\ref{fig:diagonal_ratio_plot} shows that the overall trends are consistent across a wide range of \(k\), while also highlighting clear model-dependent differences.

For SeamlessM4T, the diagonal ratio is consistently highest at the first decoding step (\(t=1\)) for all values of \(k\), demonstrating that our main finding is robust to the number of selected neurons. As generation progresses, the diagonal ratio decreases at \(t=6\) and partially recovers at \(t=11\), although this recovery is less pronounced for \(k=100\).

In contrast, Qwen2-Audio maintains an almost perfect diagonal ratio (approximately 1.0) across all values of \(k\), indicating that neurons associated with the same language remain highly consistent across modalities regardless of the number of selected neurons.

These results suggest that the strongest cross-modal alignment in SeamlessM4T is concentrated within a relatively small subset of highly language-selective neurons, whereas Qwen2-Audio exhibits robust language-specific alignment across a substantially larger fraction of the network. Consequently, increasing \(k\) preserves clear language separation in Qwen2-Audio, while introducing progressively less cross-language overlap in SeamlessM4T.

\subsection{Cross-modal effect ratio}
\label{app:ratio}
\begin{table}[t]
\centering
\small
\begin{tabular}{lllrrr}
\toprule
Task & Intervention & $t$ & ASR & TR & Ratio \\
\midrule
\multicolumn{6}{c}{$t=6$} \\
\midrule
Chinese & French   & 6  & 51.43 & 15.09 & 0.29 \\
Chinese & German   & 6  & 91.69 & 50.47 & 0.55 \\
French  & Chinese  & 6  & 77.73 & 30.70 & 0.39 \\
French  & Japanese & 6  & 65.34 & 10.34 & 0.16 \\
German  & Chinese  & 6  & 74.96 & 16.75 & 0.22 \\
German  & Japanese & 6  & 60.85 & 14.16 & 0.23 \\
Japanese& French   & 6  & 33.64 & 11.48 & 0.34 \\
Japanese& German   & 6  & 89.28 & 63.12 & 0.71 \\
\midrule
\multicolumn{6}{c}{$t=11$} \\
\midrule
Chinese & French   & 11 & 71.60 & 31.15 & 0.44 \\
Chinese & German   & 11 & 95.58 & 58.96 & 0.62 \\
French  & Chinese  & 11 & 89.49 & 50.85 & 0.57 \\
French  & Japanese & 11 & 86.76 & 25.96 & 0.30 \\
German  & Chinese  & 11 & 85.81 & 31.67 & 0.37 \\
German  & Japanese & 11 & 83.32 & 22.89 & 0.27 \\
Japanese& French   & 11 & 43.78 & 24.03 & 0.55 \\
Japanese& German   & 11 & 95.43 & 77.08 & 0.81 \\
\bottomrule
\end{tabular}
\caption{Cross-modal effect ratios for all language pairs satisfying
$\text{ControlRate}_A > \tau$ for SeamlessM4T. Ratios increase at later decoding steps,
indicating stronger cross-modal transfer.}
\label{tab:ratio}
\end{table}

We report the \textit{cross-modal effect ratio} for SeamlessM4T defined in Section~\ref{sec:method}, 
computed as $\text{ControlRate}_B / \text{ControlRate}_A$, where $A$ 
denotes the source modality (ASR) and $B$ the target modality (text repetition). 
This ratio quantifies how strongly language-control effects identified in one modality transfer to another.

To ensure reliability, we only compute ratios for cases with $\text{ControlRate}_A > \tau$, 
where $\tau = 0.1$, excluding cases where control in the source modality is too weak to yield meaningful comparisons.

Table~\ref{tab:ratio} reports cross-modal effect ratios for all valid language pairs. 
We find that the ratio consistently increases from $t=6$ to $t=11$, indicating that cross-modal transfer of language-control effects becomes stronger as generation proceeds.

While variability exists across language pairs, this trend is consistent and not driven by a small number of outliers.

\subsection{FastText-based language control evaluation}
\label{appendix:fasttext}

\begin{table*}[t]
\centering
\small
\setlength{\tabcolsep}{5pt}
\begin{tabular}{lllrrrrrrrr}
\toprule
\multirow{2}{*}{Task lang}
& \multirow{2}{*}{Task}
& \multirow{2}{*}{Intervention}
& \multicolumn{2}{c}{$t=1$}
& \multicolumn{2}{c}{$t=6$}
& \multicolumn{2}{c}{$t=11$}
& \multicolumn{2}{c}{Baseline} \\
\cmidrule(lr){4-5}
\cmidrule(lr){6-7}
\cmidrule(lr){8-9}
\cmidrule(lr){10-11}
& & & src & tgt & src & tgt & src & tgt & src & tgt \\
\midrule
German & ASR & Japanese
& 99.06 & \textbf{0.01}
& 39.63 & \textbf{40.22}
& 8.42 & \textbf{70.88}
& 99.07 & \textbf{0.01} \\

German & ASR & Chinese
& 99.07 & \textbf{0.00}
& 24.25 & \textbf{52.07}
& 3.12 & \textbf{73.29}
& 99.10 & \textbf{0.00} \\

German & TR & Japanese
& 99.09 & \textbf{0.00}
& 92.81 & \textbf{5.42}
& 83.75 & \textbf{11.86}
& 99.09 & \textbf{0.00} \\

German & TR & Chinese
& 99.10 & \textbf{0.00}
& 93.27 & \textbf{4.69}
& 84.44 & \textbf{11.04}
& 99.08 & \textbf{0.00} \\
\midrule

French & ASR & Japanese
& 98.72 & \textbf{0.01}
& 37.28 & \textbf{22.14}
& 11.20 & \textbf{39.75}
& 98.76 & \textbf{0.01} \\

French & ASR & Chinese
& 98.73 & \textbf{0.00}
& 20.09 & \textbf{41.94}
& 6.62 & \textbf{58.17}
& 98.75 & \textbf{0.01} \\

French & TR & Japanese
& 98.60 & \textbf{0.01}
& 94.21 & \textbf{1.66}
& 81.10 & \textbf{12.41}
& 98.71 & \textbf{0.01} \\

French & TR & Chinese
& 98.74 & \textbf{0.01}
& 82.32 & \textbf{13.24}
& 63.13 & \textbf{29.57}
& 98.70 & \textbf{0.01} \\
\midrule

Japanese & ASR & German
& 99.97 & \textbf{0.00}
& 15.44 & \textbf{60.18}
& 5.28 & \textbf{72.91}
& 99.97 & \textbf{0.00} \\

Japanese & ASR & French
& 99.97 & \textbf{0.00}
& 84.34 & \textbf{9.37}
& 74.87 & \textbf{14.29}
& 99.97 & \textbf{0.00} \\

Japanese & TR & German
& 99.99 & \textbf{0.00}
& 44.90 & \textbf{39.89}
& 25.47 & \textbf{54.31}
& 98.98 & \textbf{0.00} \\

Japanese & TR & French
& 98.99 & \textbf{0.00}
& 95.38 & \textbf{2.87}
& 84.50 & \textbf{6.67}
& 98.94 & \textbf{0.00} \\
\midrule

Chinese & ASR & German
& 97.52 & \textbf{0.08}
& 8.80 & \textbf{69.30}
& 1.46 & \textbf{74.73}
& 96.80 & \textbf{0.10} \\

Chinese & ASR & French
& 95.67 & \textbf{0.46}
& 52.74 & \textbf{38.05}
& 35.96 & \textbf{54.35}
& 95.98 & \textbf{0.60} \\

Chinese & TR & German
& 99.11 & \textbf{0.00}
& 63.69 & \textbf{32.26}
& 55.08 & \textbf{38.54}
& 99.09 & \textbf{0.00} \\

Chinese & TR & French
& 99.10 & \textbf{0.00}
& 91.47 & \textbf{7.51}
& 82.16 & \textbf{15.83}
& 99.16 & \textbf{0.00} \\
\bottomrule
\end{tabular}

\caption{
FastText-based evaluation of language control under neuron interventions for SeamlessM4T.
Target-language probabilities increase substantially at later decoding steps
($t=6$, $t=11$), confirming that language control becomes effective over time
and supporting the script-based results with an independent metric.
The Baseline columns report the source- and target-language probabilities
obtained by intervening on mid-ranked neurons at $t=11$.
Target-language probabilities are highlighted in bold.
}
\label{tab:unimodal_neuron_probs}
\end{table*}


To validate our script-based evaluation with an independent metric, 
we measure language probabilities using fastText for SeamlessM4T.

Table~\ref{tab:unimodal_neuron_probs} shows that target-language probabilities increase substantially at later decoding steps, 
confirming that language control becomes effective over time.

At $t=1$, the probability assigned to the target language remains close to zero across all conditions, 
indicating minimal language shift. At later decoding steps ($t=6$ and $t=11$), the target language probability increases substantially, 
and in some cases exceeds that of the source language (e.g., German$\rightarrow$Chinese in ASR at $t=6$, and Japanese$\rightarrow$German in text repetition at $t=11$).

These results are consistent with the script-based metric and provide complementary evidence that output-language control strengthens at later decoding steps. 
The fact that the same neurons induce similar shifts in both ASR and text repetition further suggests that output-language control partially relies on shared circuitry across modalities.

\subsection{Qualitative evaluation on language control}
\label{appendix:qualitative}

Interestingly, in the ASR example (Table~\ref{tab:de-zh-intervention-example}), the output shifts to Chinese lexical items while partially preserving German word order. In particular, in the gold German sentence, the verb \textit{hören} (`hear') appears at the end of the clause because it is governed by the auxiliary verb \textit{werden} (`will'). While in Chinese the verb \zh{听} (`hear') would normally be placed following the auxiliary verb \zh{会} (`will'), the model placed \zh{听} (`hear') at the end of the sentence, partially preserving German syntax.

To complement the qualitative evaluation in the main text, we present additional qualitative examples from SeamlessM4T. We manually looked into all the examples to confirm that the intervention changes the output language, and not only the script. All examples shown here are taken from $t=11$. Across these cases, the intervention often changes lexical language identity, but does not reliably impose well-formed target-language syntax.

\noindent\textbf{Example 2: Japanese $\rightarrow$ German intervention.}
\begin{tcolorbox}[examplebox]
\textbf{Gold:} \jp{寺院の一部として ジグラットと呼ばれるピラミッド型の特殊な塔が建てられることがありました}

\textbf{ASR:} \jp{寺院の一部として}Zikr\jp{と呼ばれる}pyramidalforms spezialn tower gebet gebet gebet gebet gebet gebet \ldots

\textbf{Text repetition:} \jp{寺院の一部として ジグラットと呼ばれる}Pilaamid-Types\jp{特殊な}Taar\jp{が} erbaet\jp{されることがありました。}
\end{tcolorbox}

The intervention yields unstable outputs. In ASR, decoding collapses into repetition, while in text repetition, several items are shifted toward German-like forms but remain malformed, such as \textit{Taar} instead of \textit{Turm} and \textit{erbaet} instead of \textit{erbaut}. This suggests that the intervention biases lexical language identity, but is insufficient to ensure correct word forms or coherent syntax.

\noindent\textbf{Example 3: German $\rightarrow$ Chinese intervention.}
\begin{tcolorbox}[examplebox]
\textbf{Gold:} zeit ist auch die art und weise wie wir die dauer länge von veranstaltungen vergleichen

\textbf{ASR:} \zh{时间} ist\zh{也}theart\zh{和方式,怎么我们比较活动的持续时间}

\textbf{Text repetition:} zeit ist auch die\zh{艺术和方式} wie wir die dauer länge von veranstaltungen\zh{比较}
\end{tcolorbox}

The outputs show mixed-language sequences and inconsistent word order. For example, \textit{vergleichen} is mapped to \zh{比较}, but its position differs across ASR and text repetition. The outputs also contain lexical distortions such as \textit{theart} and the contextually inappropriate translation \zh{艺术}. This suggests that the intervention affects lexical language identity without deterministically controlling syntax.

\noindent\textbf{Example 4: Chinese $\rightarrow$ French intervention.}
\begin{tcolorbox}[examplebox]
\textbf{Gold:} \zh{11 点 20 分 警 察 要 求 抗 议 者 退 回 到 人 行 道 并 告 知 抗 议 者 在 行 使 抗 议 权 利 时 务 必 考 虑 到 越 来 越 拥 堵 的 公 共 交 通 问 题}

\textbf{ASR:} onze heures vingt police demande les protestants de revenir à la rue des gens et d'invoquer les protestants de se réveiller les protestants de se réveiller les protestants de se réveiller \ldots

\textbf{Text repetition:} \zh{11 点 20 分 警 察 要 求 抗 议 者 退 回 到 人 行 道 并 告 知 抗 议 者} En \zh{行 使 抗 议 权 利 时 务 必 考 虑} En \zh{越} venir \zh{越 拥 堵 的 公 共 交 通 问 题}
\end{tcolorbox}

The ASR output begins with plausible French material, such as \textit{onze heures vingt}, but soon degenerates into repetition. In text repetition, most of the sentence remains in Chinese, with isolated French insertions. Notably, part of the fixed expression \zh{越来 越} is translated as \textit{venir}, suggesting that the model translates subcomponents of a construction rather than preserving the construction as a unit. This again shows that the intervention affects lexical language identity more strongly than higher-level structural planning.

Across Examples 2--5, the intervention consistently influences lexical language identity, but does not reliably enforce target-language syntax. The outputs often preserve source-language structure, produce malformed target-language forms, or become unstable during decoding. These cases therefore support our interpretation that the identified language neurons contribute to output-language control, but are not sufficient to account for full syntactic planning.

\subsection{Overlap matrices}
\label{appendix:matrices}

\begin{figure*}[t]
\centering

\begin{subfigure}[t]{0.32\textwidth}
    \centering
    \includegraphics[width=\linewidth]{
        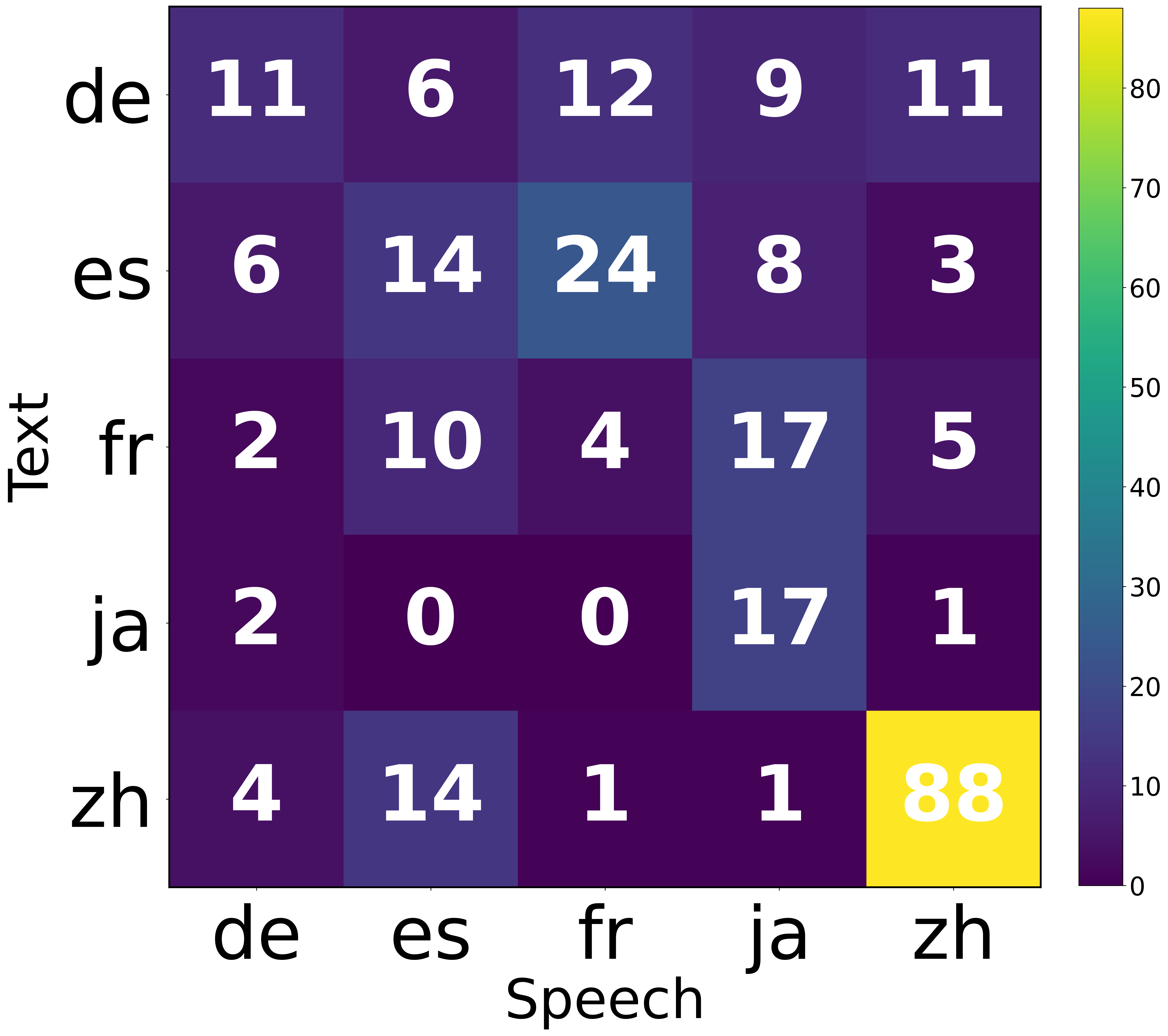
    }
    \caption{Bottom, $t=1$}
\end{subfigure}\hfill
\begin{subfigure}[t]{0.32\textwidth}
    \centering
    \includegraphics[width=\linewidth]{
        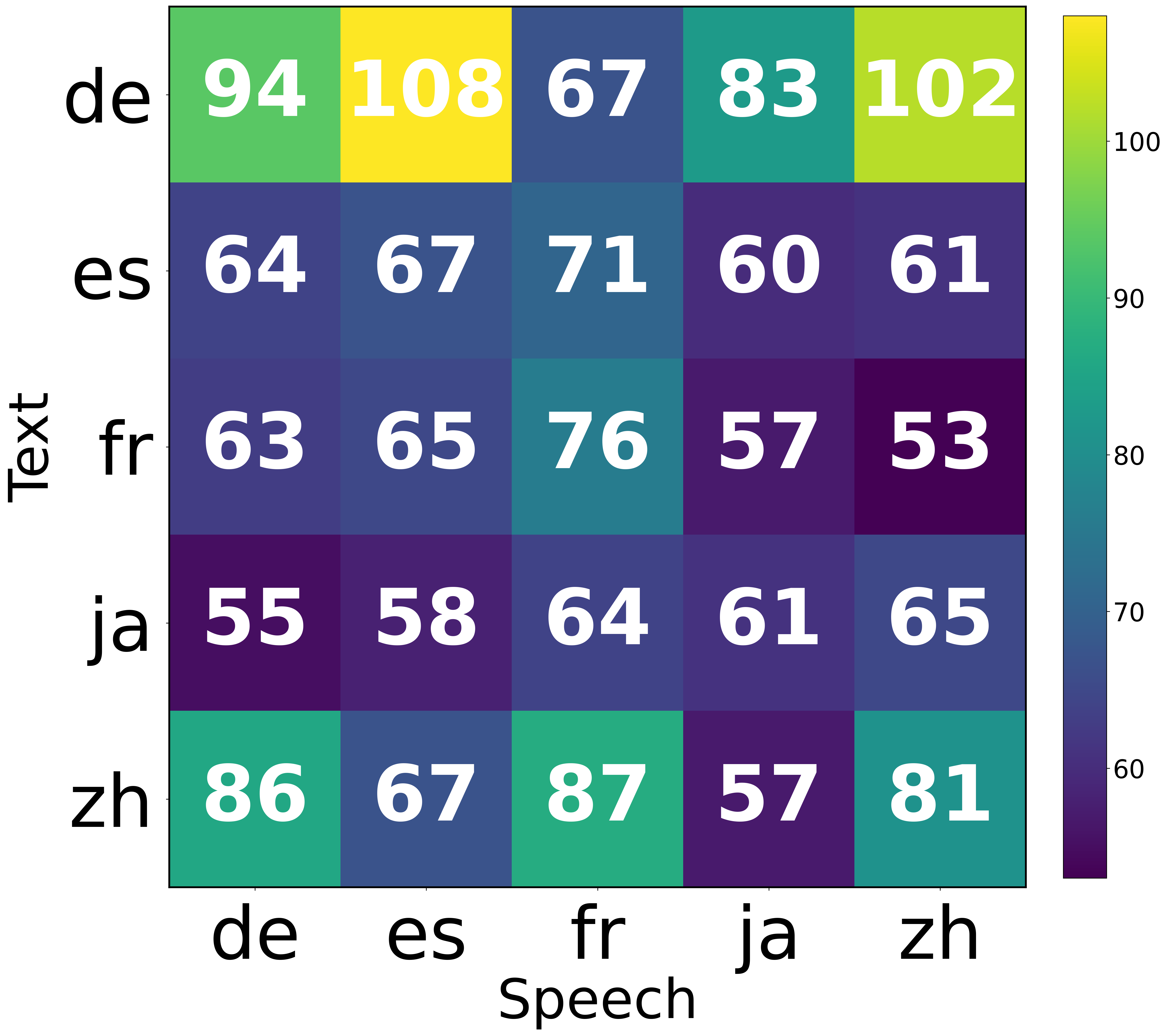
    }
    \caption{Random, $t=1$}
\end{subfigure}\hfill
\begin{subfigure}[t]{0.32\textwidth}
    \centering
    \includegraphics[width=\linewidth]{
        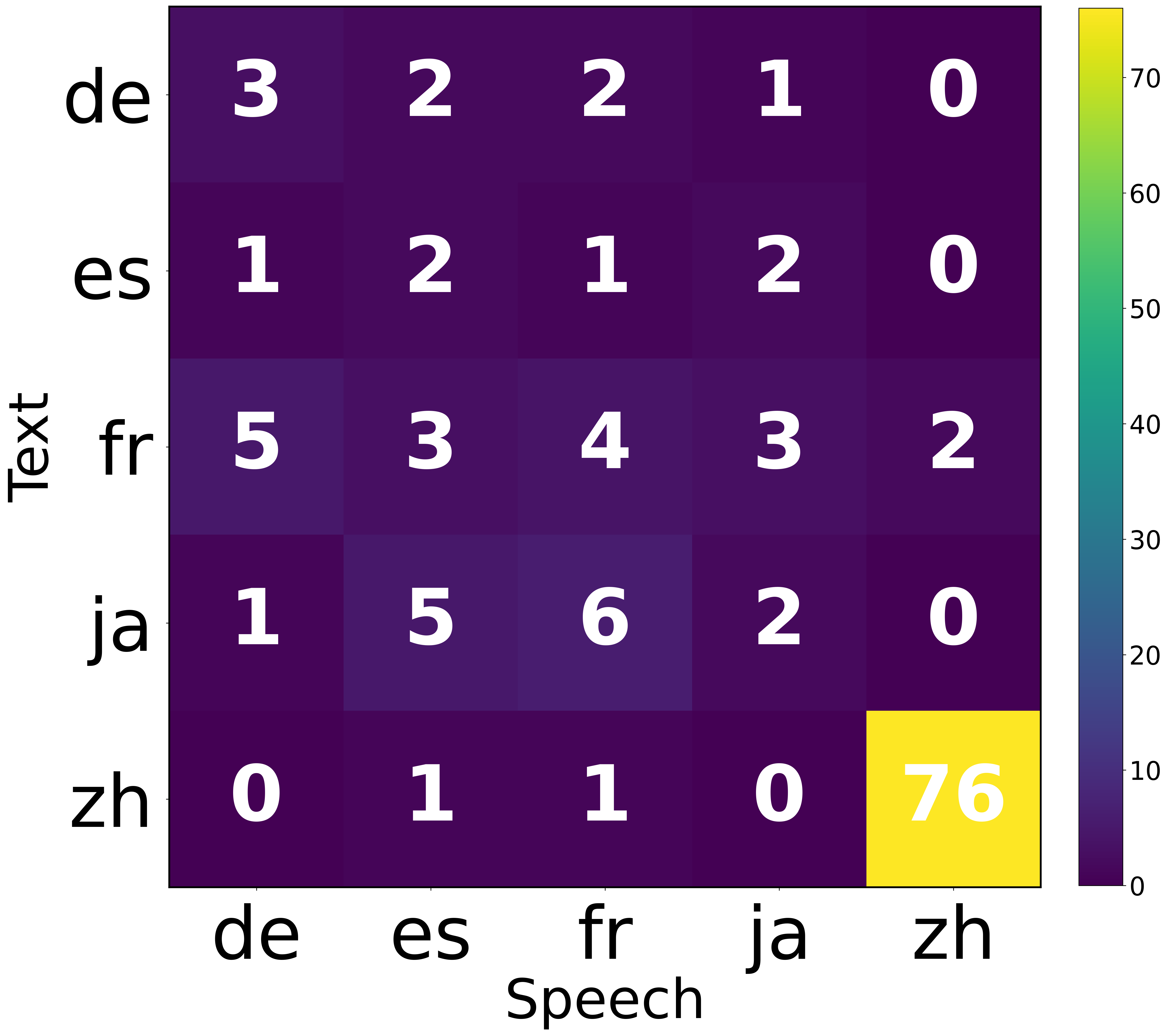
    }
    \caption{Mid, $t=1$}
\end{subfigure}

\par

\begin{subfigure}[t]{0.32\textwidth}
    \centering
    \includegraphics[width=\linewidth]{
        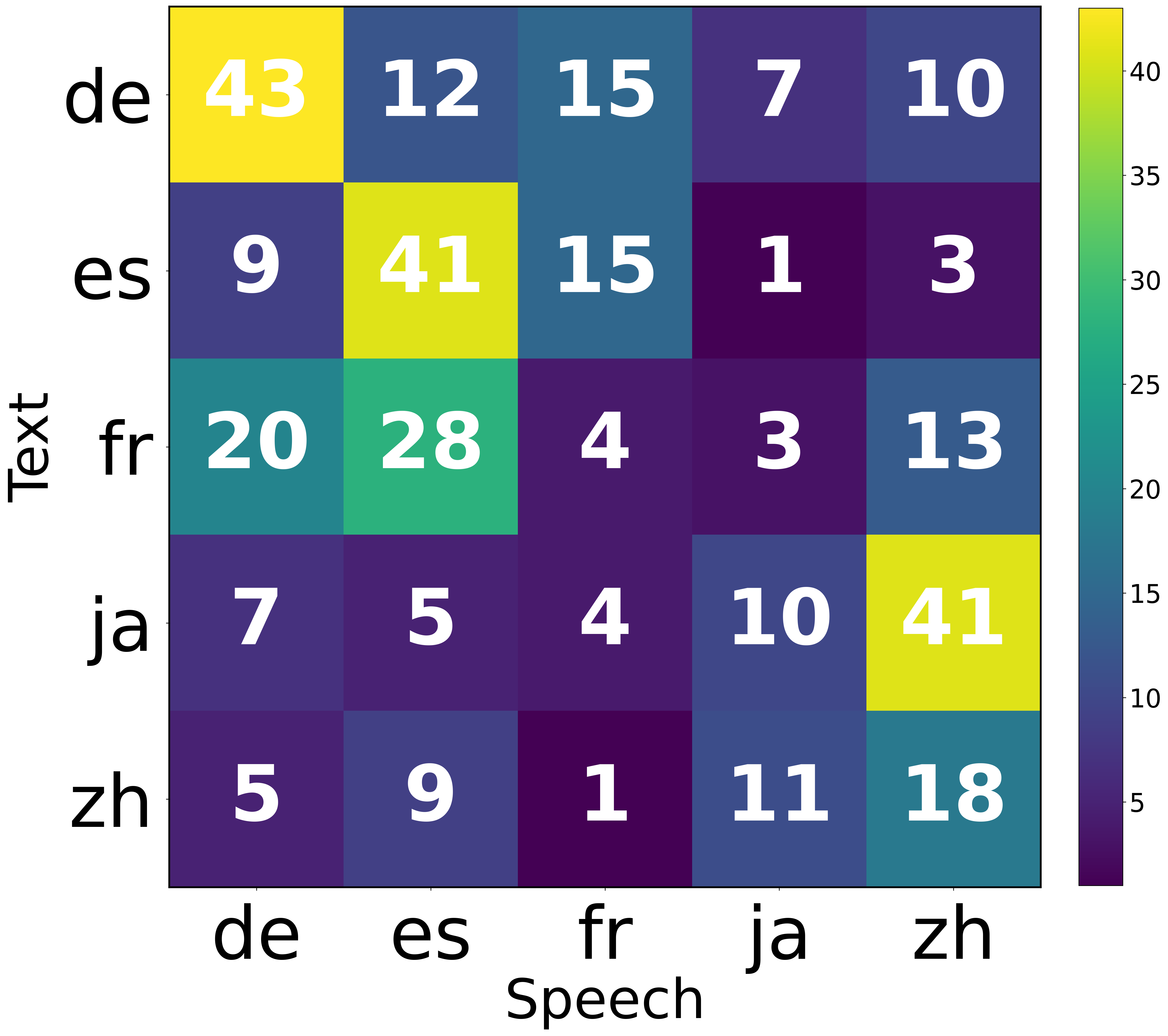
    }
    \caption{Top, $t=6$}
\end{subfigure}\hfill
\begin{subfigure}[t]{0.32\textwidth}
    \centering
    \includegraphics[width=\linewidth]{
        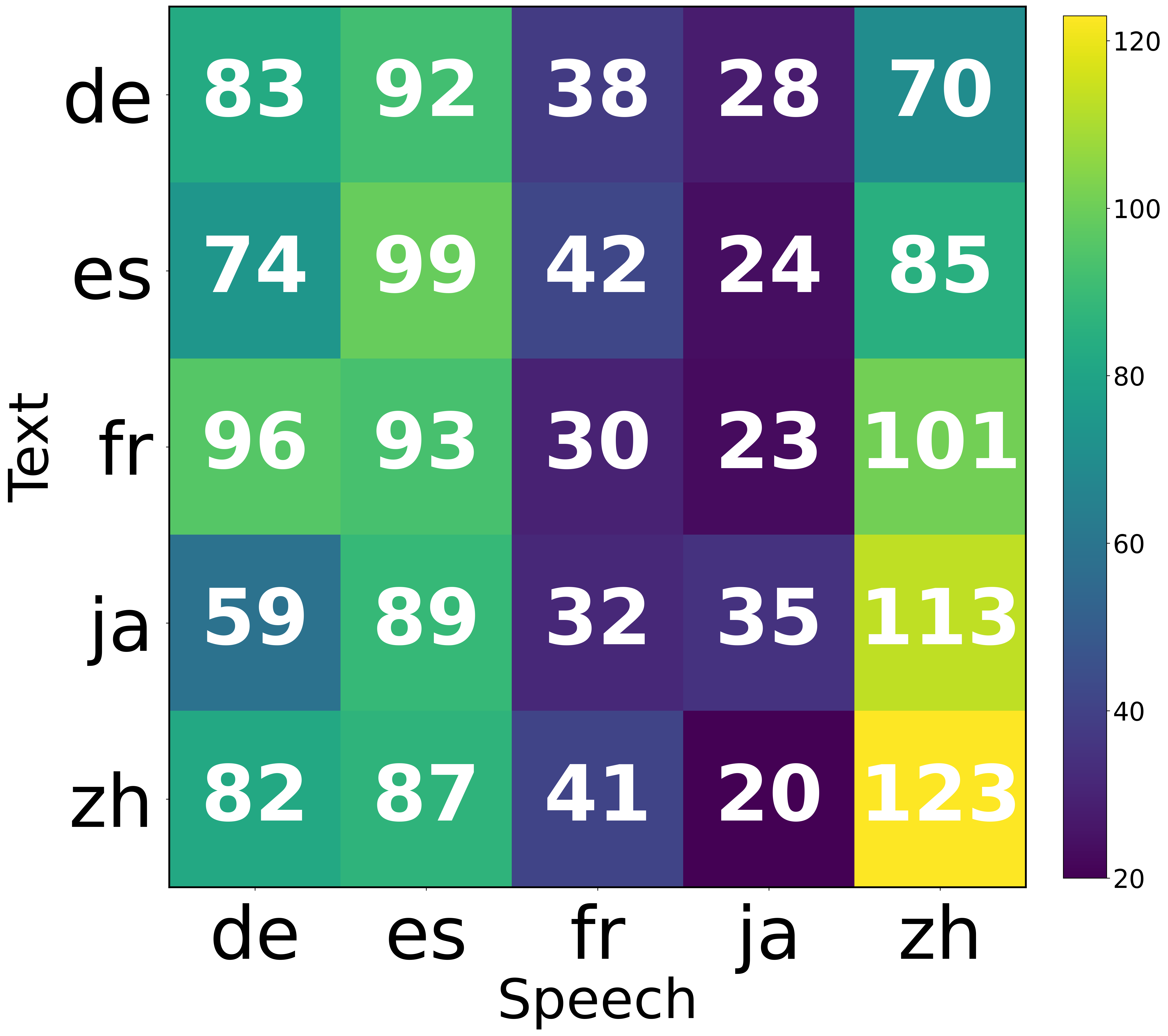
    }
    \caption{Random, $t=6$}
\end{subfigure}\hfill
\begin{subfigure}[t]{0.32\textwidth}
    \centering
    \includegraphics[width=\linewidth]{
        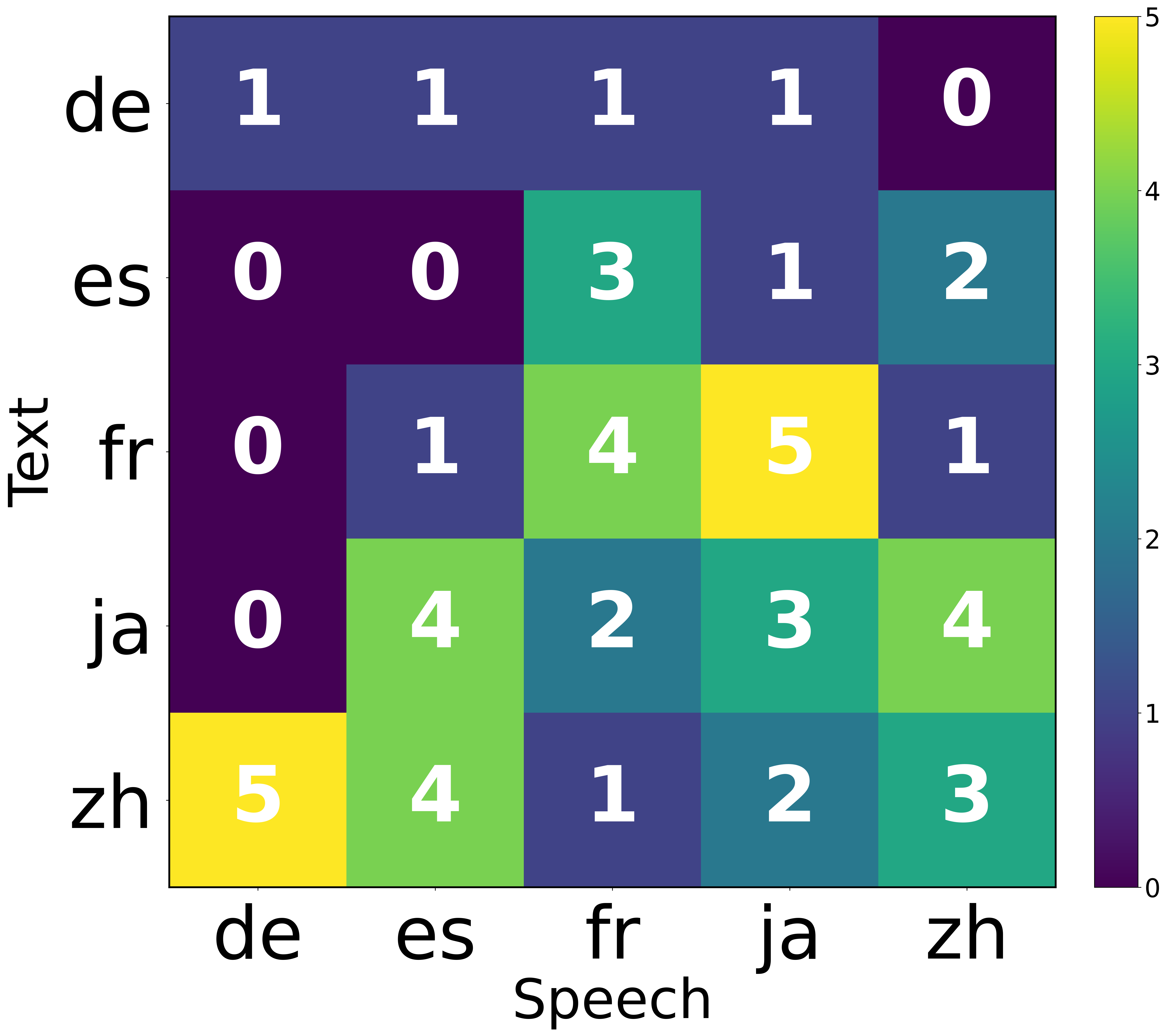
    }
    \caption{Mid, $t=6$}
\end{subfigure}

\caption{
Cross-modal neuron-overlap heatmaps for SeamlessM4T.
The top row compares bottom-ranked, random, and mid-ranked neurons at $t=1$.
The bottom row compares top-ranked, random, and mid-ranked neurons at $t=6$.
}
\label{fig:overlap_heatmaps_seamless_baselines}
\end{figure*}

\begin{figure*}[t]
\centering

\begin{subfigure}[t]{0.32\textwidth}
    \centering
    \includegraphics[width=\linewidth]{
        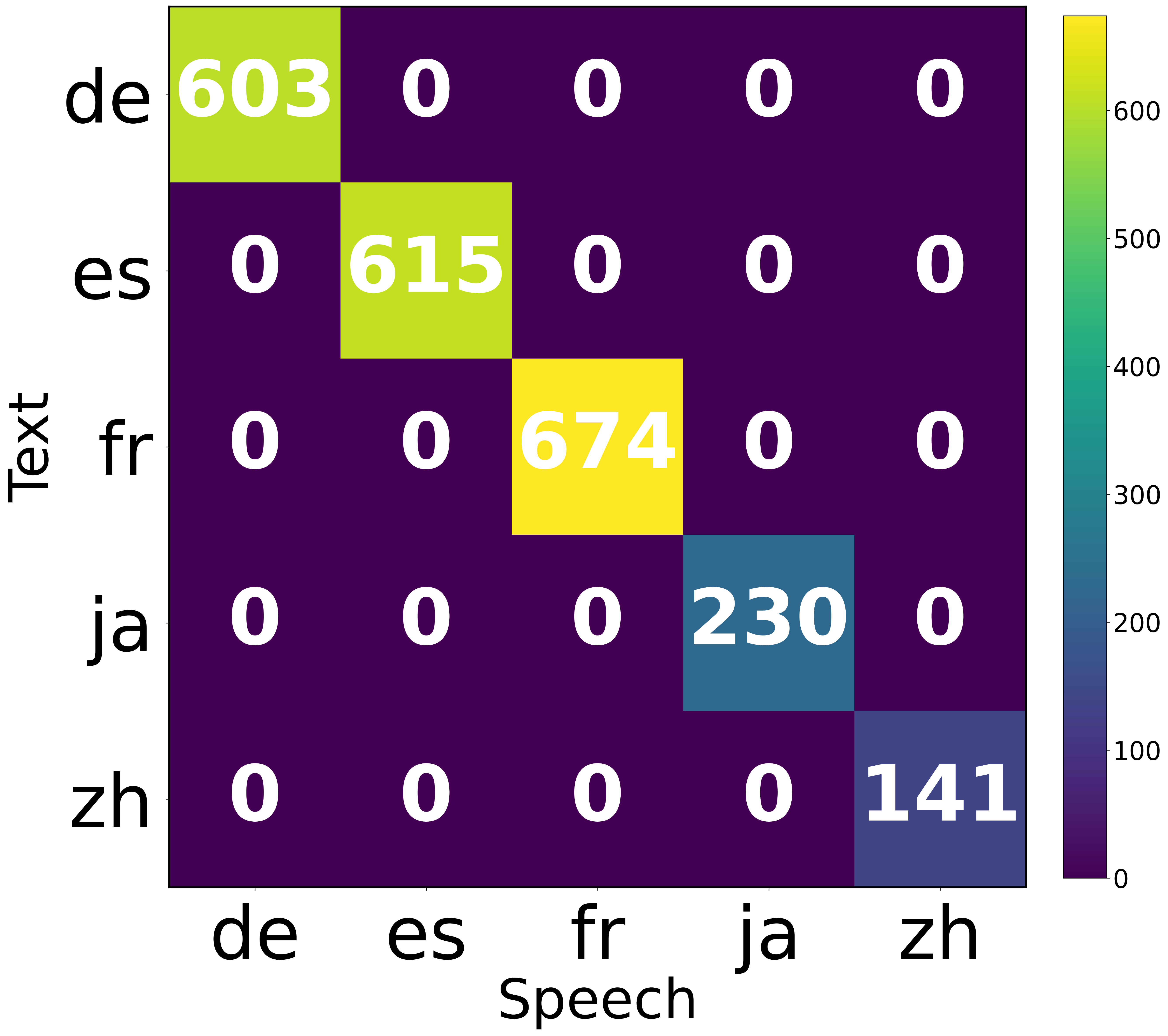
    }
    \caption{Top, $t=1$}
\end{subfigure}\hfill
\begin{subfigure}[t]{0.32\textwidth}
    \centering
    \includegraphics[width=\linewidth]{
        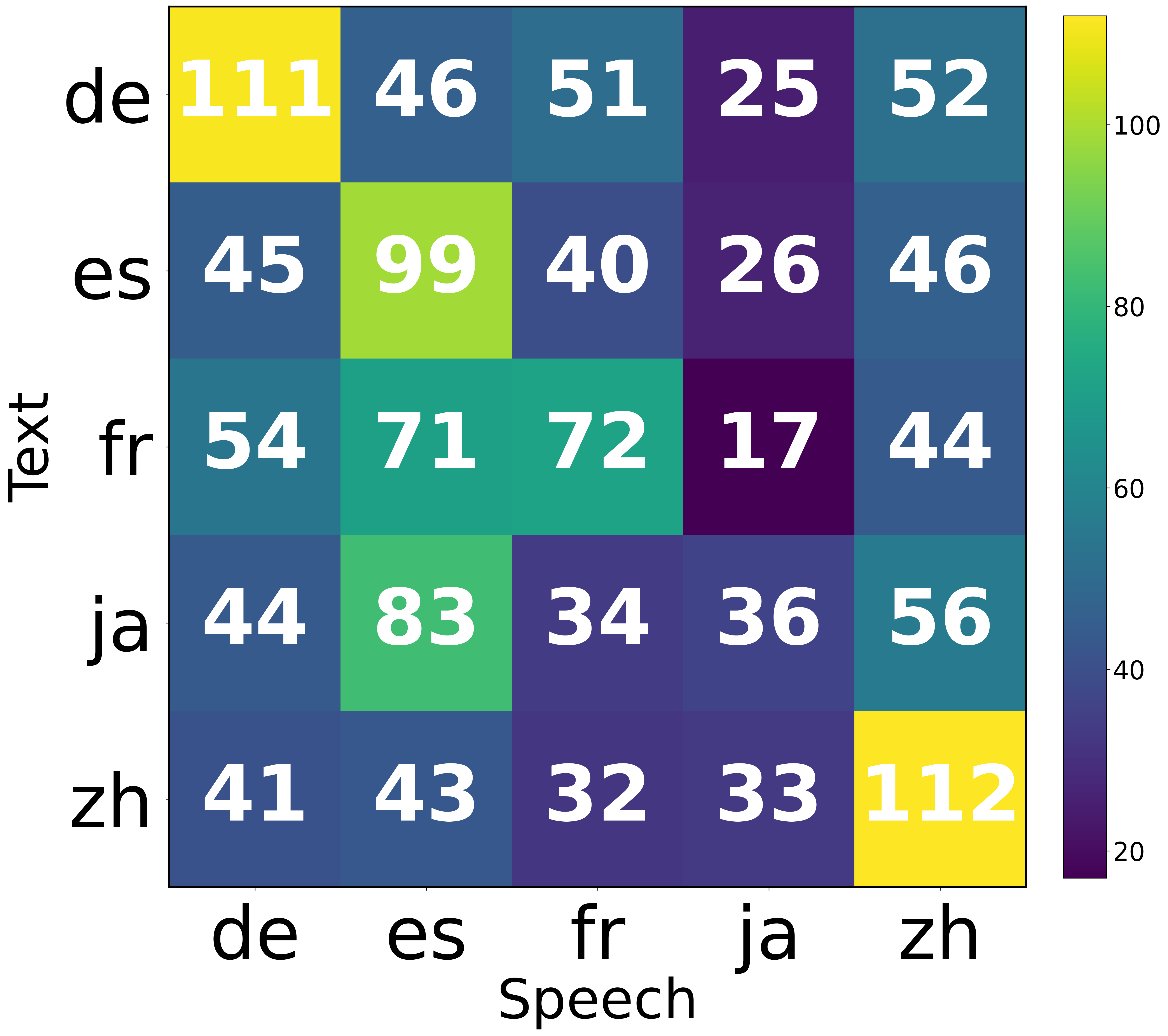
    }
    \caption{Random, $t=1$}
\end{subfigure}\hfill
\begin{subfigure}[t]{0.32\textwidth}
    \centering
    \includegraphics[width=\linewidth]{
        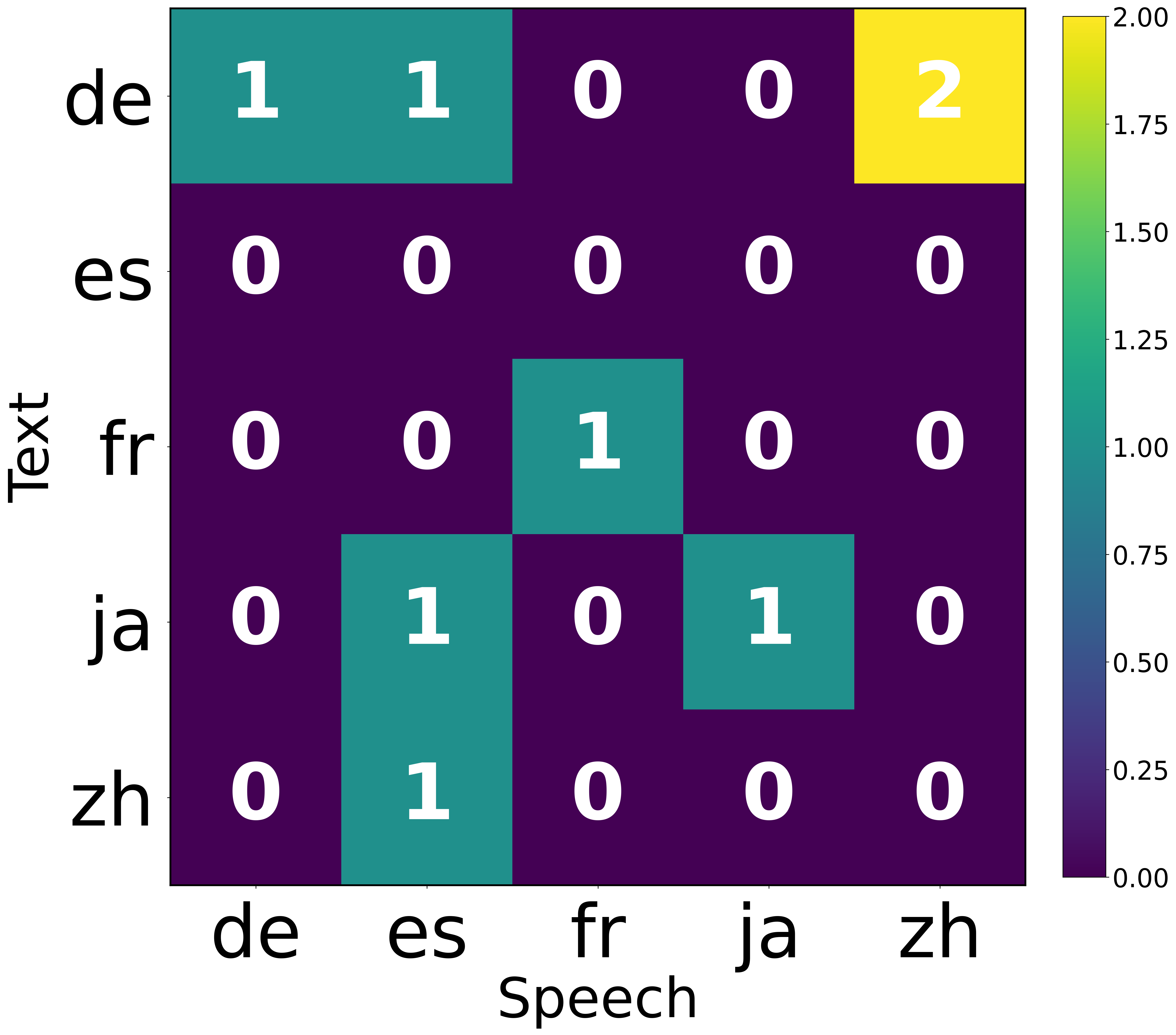
    }
    \caption{Mid, $t=1$}
\end{subfigure}

\par

\begin{subfigure}[t]{0.32\textwidth}
    \centering
    \includegraphics[width=\linewidth]{
        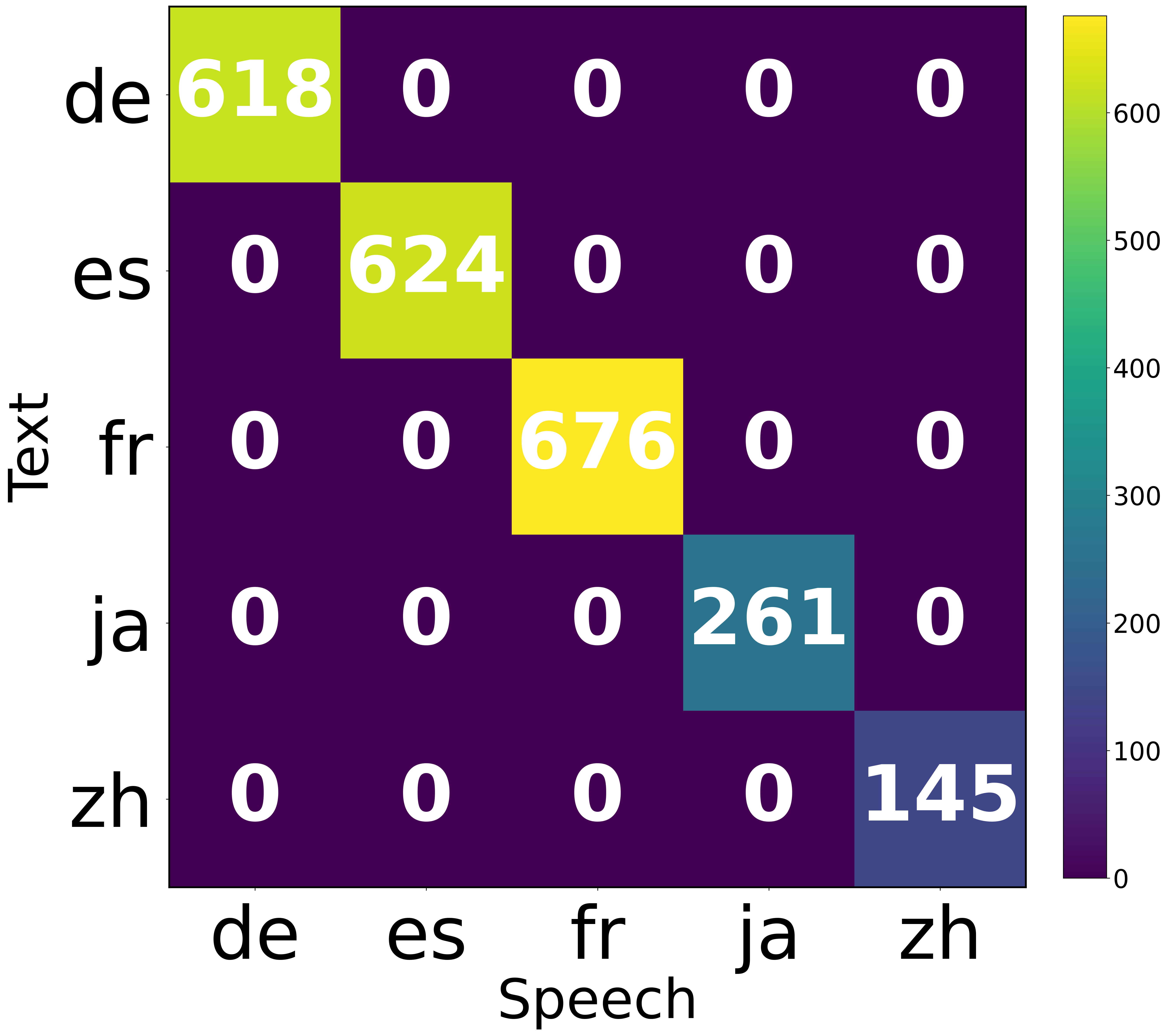
    }
    \caption{Top, $t=6$}
\end{subfigure}\hfill
\begin{subfigure}[t]{0.32\textwidth}
    \centering
    \includegraphics[width=\linewidth]{
        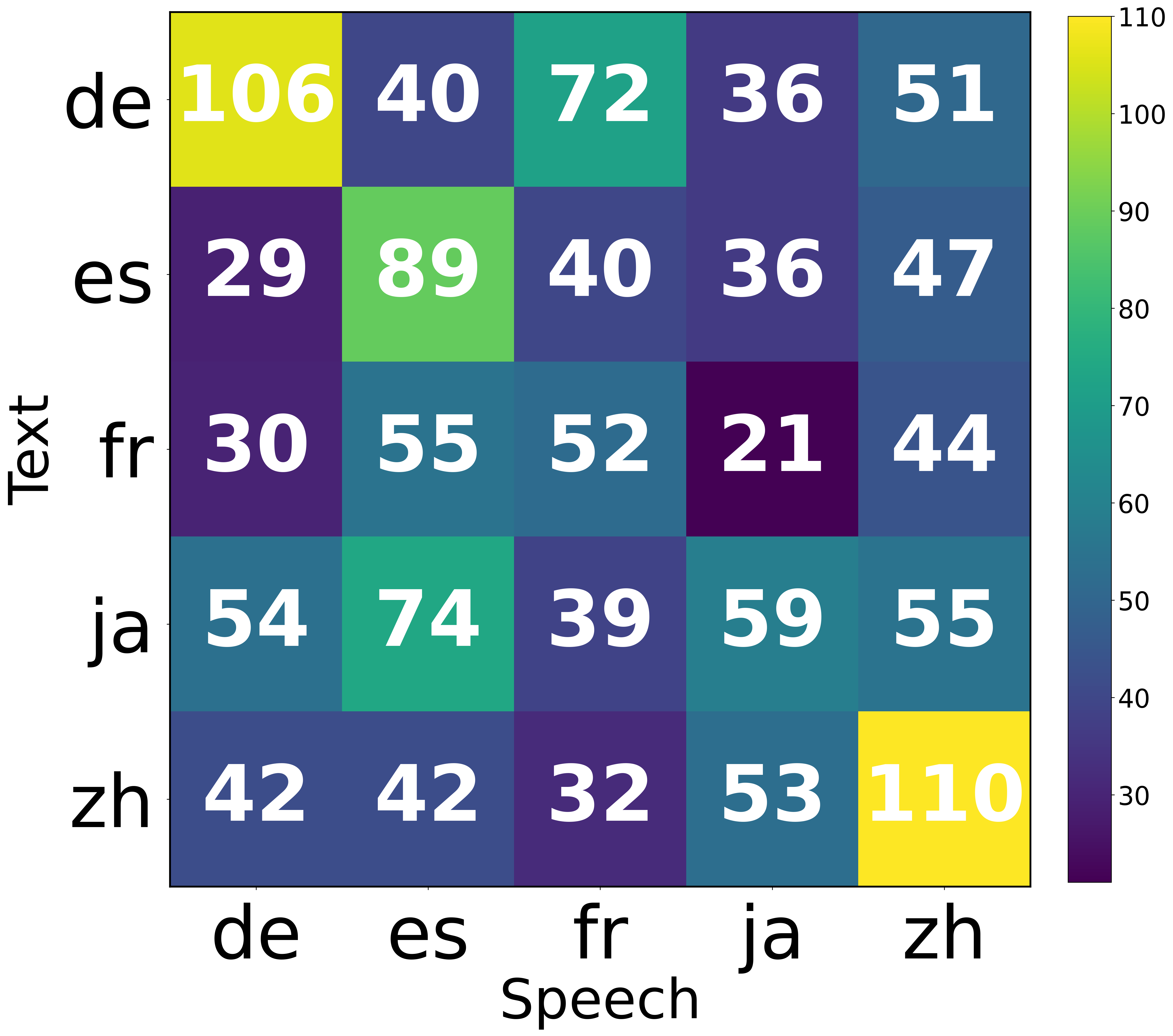
    }
    \caption{Random, $t=6$}
\end{subfigure}\hfill
\begin{subfigure}[t]{0.32\textwidth}
    \centering
    \includegraphics[width=\linewidth]{
        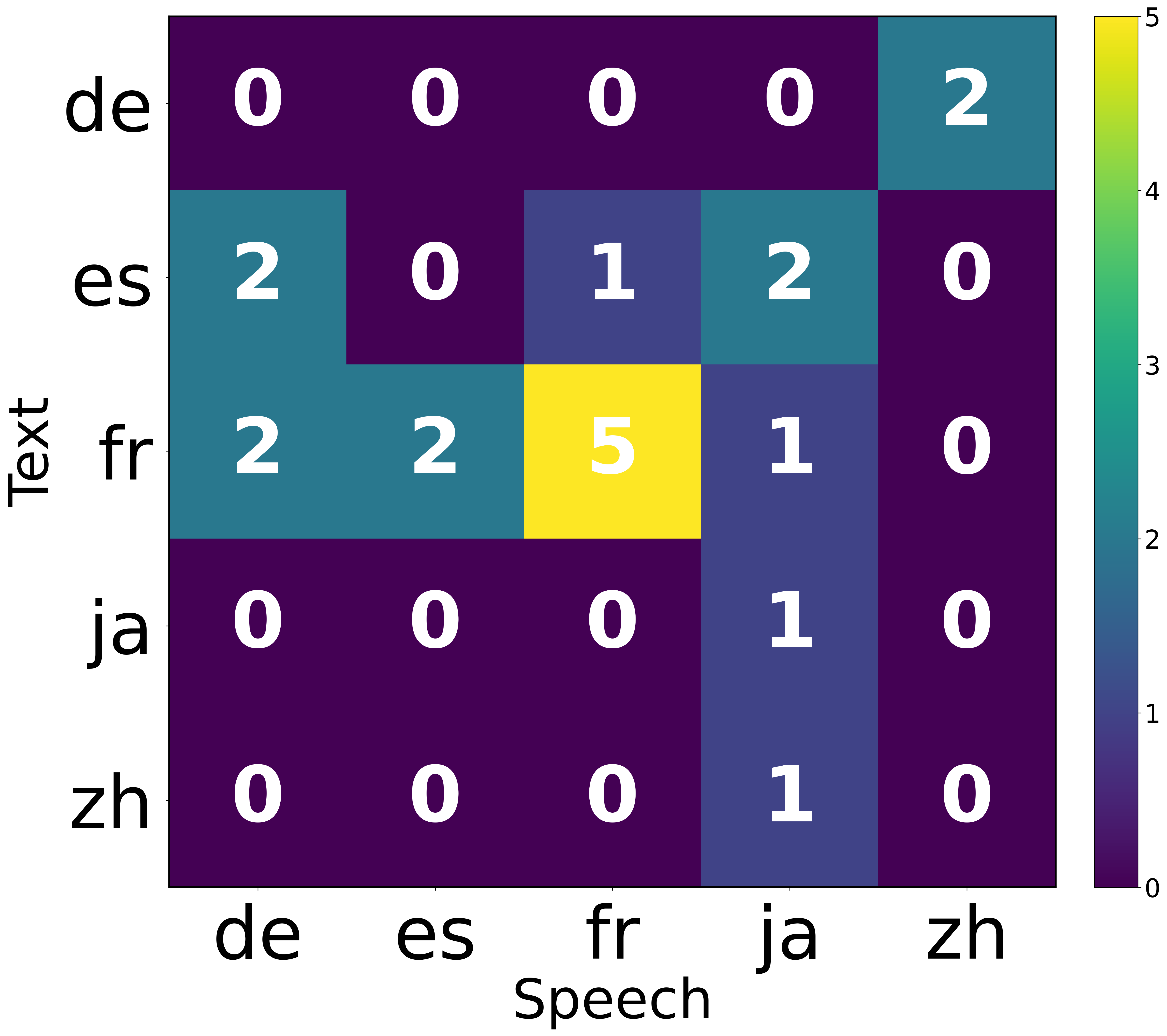
    }
    \caption{Mid, $t=6$}
\end{subfigure}

\caption{
Cross-modal neuron-overlap heatmaps for Qwen2-Audio.
The top row compares top-ranked, random, and mid-ranked neurons at $t=1$,
and the bottom row presents the corresponding comparison at $t=6$.
}
\label{fig:overlap_heatmaps_qwen_baselines}
\end{figure*}

Figures~\ref{fig:overlap_heatmaps_seamless_baselines} and~\ref{fig:overlap_heatmaps_qwen_baselines} present the complete cross-modal overlap matrices for all decoding steps and baseline methods (Random, Mid, and Ours) for SeamlessM4T and Qwen2-Audio, respectively. These figures complement the main-text analysis by illustrating the full overlap structure across decoding conditions.

\subsection{Additional decoding steps}
\label{appendix:additional-steps}

To complement the main-text analysis, we additionally evaluate intermediate decoding steps (\(t=3,5,7,9\)).

Table~\ref{tab:overlap_metrics_timesteps} reports the cross-modal representation results for both SeamlessM4T and Qwen2-Audio. For both models, the intermediate decoding steps exhibit relatively stable behavior. In particular, the representation shift observed in SeamlessM4T primarily reflects the difference between the first decoding step (\(t=1\)) and subsequent decoding, rather than a gradual change across intermediate steps.

Table~\ref{tab:intermediate_script_control} reports the cross-modal language-control results for SeamlessM4T at the additional decoding steps. As decoding progresses, both the ASR and text-repetition tasks exhibit progressively larger script shifts. Consistent with this trend, the cross-modal effect ratio remains stable at 0.31 for \(t=3\) and \(t=5\), before increasing to 0.41 at \(t=7\), 0.47 at \(t=9\), and 0.49 at \(t=11\). This steady increase suggests that neurons identified at later decoding steps become progressively more effective for cross-modal language control.

\subsection{A.11 Additional evaluation on Qwen2-Audio (Language control)}
\label{appendix:language-control-qwen}

Table~\ref{tab:qwen_en2x_unimodal_neuron_scripts} presents the results of our cross-modal language-control experiments on Qwen2-Audio. Overall, interventions from Latin-script languages to CJK languages are more successful than those in the opposite direction, although the overall transfer effect remains substantially weaker than in SeamlessM4T.

\begin{tcolorbox}[examplebox]
\textbf{Gold:}
in einem fremden land zu leben und einen freiwilligendienst zu absolvieren ist eine tolle möglichkeit eine andere kultur kennenzulernen neue menschen zu treffen etwas über sich selbst zu lernen die dinge aus einer anderen perspektive zu sehen und sogar neue fähigkeiten zu erlernen

\textbf{ASR:}
\zh{毫无疑义,在陌生的国家里生活和做义务服务是一种极好的机会,可以了解另一种文化,认识新的人,学习新的东西,从不同的角度观察事物,甚至获得新的能力.}

\textbf{Text repetition:}
\zh{但不限制。}
\end{tcolorbox}

The ASR output is fluent Chinese and remains semantically faithful to the source sentence, demonstrating successful language control. In contrast, the text-repetition output immediately degenerates into an unrelated Chinese phrase, indicating that the identified neurons do not reliably transfer across modalities for this example.

\begin{tcolorbox}[examplebox]
\textbf{Gold:}
\jp{クロアチアの首都ザグレブで生まれたボベックは パルチザンベオグラードで活躍していた頃に有名になりました}

\textbf{ASR:}
Kroatiens hovedstadsgruppe\jp{で生まれた}Wobek\jp{は}Patsan Beo Grat\jp{で活躍していた頃に有名になりました}

\textbf{Text repetition:}
\jp{ザグレブ出身のボベックは、パルチザンベオグラードで活躍した頃に有名になりました。}
 Korean: Bojicic\ko{는} Zagreb\ko{에서 태어났고}, Parizan Beograd\ko{에서 활약했을 때 유명해졌습니다.}
\end{tcolorbox}

This example illustrates a failure case for CJK-to-Latin transfer. Although the ASR output begins with a German word, it also contains the Danish word \emph{hovedstads} and otherwise largely preserves the original Japanese sentence, indicating incomplete language switching. Likewise, the text-repetition output first continues in Japanese before switching to Korean. Such mixed-language outputs expose a limitation of our script-based evaluation metric, as script-level changes alone do not guarantee successful transfer to the intended target language. This observation highlights the importance of comparing our method with the Random and Mid baselines.

Overall, the qualitative examples are consistent with the quantitative results. Latin-to-CJK interventions are generally more successful than CJK-to-Latin interventions, and a weak temporal trend is observed, with slightly better performance at \(t=1\) than at later decoding steps. However, this temporal effect is modest and warrants further investigation.

\subsection{Additional implementation details}


\paragraph{Speech preprocessing.}
All speech inputs are resampled to 16 kHz. Log-mel filterbank features are extracted with \( n_{\text{mels}} = 160 \), and a fixed temporal length of \( T = 300 \) is enforced via truncation or zero-padding.

\paragraph{Subcomponents.}
For Conformer components with activations of shape \( d \times n \), we consistently treat the hidden dimension \( d \) as the neuron axis.

We exclude attention distance-embedding activations from our experiment, as they do not possess a neuron (hidden) dimension.

\begin{table*}[t]
\centering
\small
\setlength{\tabcolsep}{6pt}
\renewcommand{\arraystretch}{1.15}

\begin{tabular}{llcccc}
\toprule
Model & Metric & $t=3$ & $t=5$ & $t=7$ & $t=9$ \\
\midrule

\multirow{4}{*}{Seamless}
& Cross-modal share
& 0.06 & \textbf{0.07} & 0.07 & 0.07 \\

& Same-language
& \textbf{0.21} & 0.20 & 0.20 & 0.21 \\

& Correct typological
& \textbf{0.18} & 0.17 & 0.18 & 0.17 \\

& Wrong typological
& \textbf{-0.55} & -0.53 & -0.55 & -0.54 \\

\midrule

\multirow{4}{*}{Qwen2-Audio}
& Cross-modal share
& 0.45 & \textbf{0.47} & 0.46 & 0.47 \\

& Same-language
& \textbf{1.00} & \textbf{1.00} & \textbf{1.00} & \textbf{1.00} \\

& Correct typological
& \textbf{-0.47} & \textbf{-0.47} & \textbf{-0.47} & \textbf{-0.47} \\

& Wrong typological
& \textbf{-0.92} & \textbf{-0.92} & \textbf{-0.92} & \textbf{-0.92} \\

\bottomrule
\end{tabular}
\caption{
Comparison of overlap metrics on additional decoding steps ($t=3,5,7,9$). Cross-modal share reports the fraction of shared language-selective neurons across modalities, while same-language, correct typological, and wrong typological overlap are chance-corrected. Both SeamlessM4T and Qwen2-Audio exhibit relatively stable overlap metrics across the intermediate decoding steps.
}
\label{tab:overlap_metrics_timesteps}
\end{table*}

\begin{table*}[t]
\centering
\small
\setlength{\tabcolsep}{5pt}
\begin{tabular}{lllrrrr}
\toprule
\multirow{2}{*}{Task lang}
& \multirow{2}{*}{Task}
& \multirow{2}{*}{Intervention}
& \multicolumn{4}{c}{Ours} \\
\cmidrule(lr){4-7}
& & & $t=3$ & $t=5$ & $t=7$ & $t=9$ \\
\midrule

German & S2T & Japanese
& \textbf{3.29} & \textbf{44.47} & \textbf{71.52} & \textbf{79.40} \\
German & S2T & Chinese
& \textbf{34.26} & \textbf{67.49} & \textbf{76.38} & \textbf{85.98} \\
German & T2T & Japanese
& \textbf{0.45} & \textbf{8.68} & \textbf{16.48} & \textbf{21.94} \\
German & T2T & Chinese
& \textbf{3.32} & \textbf{12.29} & \textbf{28.58} & \textbf{28.52} \\

\midrule

French & S2T & Japanese
& \textbf{2.50} & \textbf{55.92} & \textbf{71.71} & \textbf{84.16} \\
French & S2T & Chinese
& \textbf{26.95} & \textbf{64.78} & \textbf{86.10} & \textbf{90.74} \\
French & T2T & Japanese
& \textbf{2.42} & \textbf{6.28} & \textbf{12.91} & \textbf{21.23} \\
French & T2T & Chinese
& \textbf{4.25} & \textbf{23.14} & \textbf{39.86} & \textbf{49.16} \\

\midrule

Japanese & S2T & German
& \textbf{52.19} & \textbf{86.67} & \textbf{92.99} & \textbf{94.87} \\
Japanese & S2T & French
& \textbf{4.45} & \textbf{20.06} & \textbf{33.90} & \textbf{42.39} \\
Japanese & T2T & German
& \textbf{16.70} & \textbf{54.52} & \textbf{67.13} & \textbf{71.61} \\
Japanese & T2T & French
& \textbf{1.10} & \textbf{3.62} & \textbf{14.77} & \textbf{23.50} \\

\midrule

Chinese & S2T & German
& \textbf{55.64} & \textbf{87.16} & \textbf{93.47} & \textbf{94.60} \\
Chinese & S2T & French
& \textbf{13.35} & \textbf{39.84} & \textbf{64.61} & \textbf{77.33} \\
Chinese & T2T & German
& \textbf{14.14} & \textbf{41.66} & \textbf{50.00} & \textbf{57.52} \\
Chinese & T2T & French
& \textbf{3.65} & \textbf{13.17} & \textbf{22.28} & \textbf{32.43} \\

\bottomrule
\end{tabular}

\caption{
Target-script rate (\%) under language-control-neuron interventions at
intermediate decoding steps on SeamlessM4T.
For interventions toward Japanese or Chinese, the table reports the CJK-script
rate; for interventions toward German or French, it reports the Latin-script
rate.
}
\label{tab:intermediate_script_control}
\end{table*}

\begin{table*}[t]
\centering
\small
\setlength{\tabcolsep}{3.5pt}
\begin{tabular}{lllrrrrrrrrr}
\toprule
\multirow{2}{*}{Task lang}
& \multirow{2}{*}{Task}
& \multirow{2}{*}{Intervention}
& \multicolumn{3}{c}{$t=1$}
& \multicolumn{3}{c}{$t=6$}
& \multicolumn{3}{c}{$t=11$} \\
\cmidrule(lr){4-6}
\cmidrule(lr){7-9}
\cmidrule(lr){10-12}
& & &
Random & Mid & Ours &
Random & Mid & Ours &
Random & Mid & Ours \\
\midrule

German & ASR & Japanese
& 0.07 & 0.06 & \textbf{16.93}
& 0.06 & 0.16 & \textbf{0.07}
& 0.07 & 0.07 & \textbf{0.07} \\

German & ASR & Chinese
& 0.07 & 0.07 & \textbf{29.85}
& 0.07 & 0.07 & \textbf{32.41}
& 0.06 & 0.07 & \textbf{26.46} \\

German & TR & Japanese
& 0.44 & 0.00 & \textbf{12.63}
& 0.00 & 0.00 & \textbf{0.00}
& 0.00 & 0.00 & \textbf{0.26} \\

German & TR & Chinese
& 0.00 & 0.00 & \textbf{13.80}
& 0.00 & 0.00 & \textbf{14.02}
& 0.00 & 0.00 & \textbf{25.22} \\

\midrule

French & ASR & Japanese
& 0.46 & 0.39 & \textbf{35.40}
& 0.35 & 0.43 & \textbf{0.63}
& 0.00 & 0.45 & \textbf{0.62} \\

French & ASR & Chinese
& 0.43 & 0.46 & \textbf{22.10}
& 0.47 & 0.46 & \textbf{33.00}
& 0.46 & 0.44 & \textbf{36.84} \\

French & TR & Japanese
& 0.00 & 0.00 & \textbf{29.97}
& 0.45 & 0.06 & \textbf{0.06}
& 0.00 & 0.00 & \textbf{0.06} \\

French & TR & Chinese
& 0.00 & 0.06 & \textbf{17.36}
& 0.00 & 0.06 & \textbf{20.92}
& 0.03 & 0.06 & \textbf{28.70} \\

\midrule

Japanese & ASR & German
& 0.07 & 0.20 & \textbf{5.43}
& 0.31 & 0.79 & \textbf{4.49}
& 0.13 & 0.43 & \textbf{4.95} \\

Japanese & ASR & French
& 0.07 & 0.13 & \textbf{0.07}
& 0.37 & 0.13 & \textbf{0.07}
& 0.80 & 0.07 & \textbf{0.07} \\

Japanese & TR & German
& 45.51 & 44.82 & \textbf{33.72}
& 42.59 & 41.25 & \textbf{35.75}
& 46.59 & 36.06 & \textbf{34.86} \\

Japanese & TR & French
& 43.50 & 42.03 & \textbf{40.61}
& 44.53 & 44.04 & \textbf{42.67}
& 41.63 & 44.91 & \textbf{42.70} \\

\midrule

Chinese & ASR & German
& 10.14 & 10.30 & \textbf{5.46}
& 10.40 & 13.61 & \textbf{8.61}
& 10.11 & 4.54 & \textbf{7.51} \\

Chinese & ASR & French
& 7.62 & 14.04 & \textbf{16.84}
& 14.41 & 4.82 & \textbf{18.75}
& 8.80 & 15.42 & \textbf{15.34} \\

Chinese & TR & German
& 28.86 & 16.52 & \textbf{27.51}
& 37.19 & 21.90 & \textbf{23.73}
& 34.98 & 28.89 & \textbf{24.94} \\

Chinese & TR & French
& 35.32 & 27.04 & \textbf{32.49}
& 32.48 & 20.15 & \textbf{32.31}
& 25.27 & 34.40 & \textbf{33.50} \\

\bottomrule
\end{tabular}

\caption{
Target-script rate (\%) under random-neuron, mid-ranked-neuron, and
language-control-neuron interventions on Qwen2-Audio.
For German and French task outputs, the table reports the CJK-script rate
(Latin-to-CJK control); for Japanese and Chinese task outputs, it reports the
Latin-script rate (CJK-to-Latin control).
\textit{Random} denotes intervention on randomly selected neurons matched with layers and subcomponents,
\textit{Mid} denotes intervention on mid-ranked neurons, and
\textit{Ours} denotes intervention on the identified language-control neurons.
Values for our method are highlighted in bold.
}
\label{tab:qwen_en2x_unimodal_neuron_scripts}
\end{table*}

\end{document}